\documentclass[10pt, a4paper]{dukenlp_tempalte/dukenlp}

\usepackage[utf8]{inputenc}
\usepackage[T1]{fontenc}
\usepackage{hyperref}
\usepackage{url}
\usepackage{fontawesome5}
\usepackage{natbib}
\usepackage{booktabs}
\usepackage{amsfonts}
\usepackage{nicefrac}
\usepackage{microtype}
\usepackage{xcolor}
\usepackage{tabularx}
\usepackage{multirow}
\usepackage{longtable}
\usepackage[most]{tcolorbox}
\usepackage{listings}
\tcbuselibrary{listings}
\usepackage{wrapfig}
\usepackage{enumitem}
\usepackage{cleveref}

\definecolor{rubricD}{HTML}{2E6FBA}
\definecolor{rubricC}{HTML}{B23A48}
\definecolor{rubricM}{HTML}{2E8B57}
\definecolor{rubricQ}{HTML}{6A4C93}
\definecolor{rubricR}{HTML}{D97706}
\definecolor{rubricI}{HTML}{0F766E}
\definecolor{rubricA}{HTML}{1F4E8C}
\definecolor{rubricE}{HTML}{B45309}
\definecolor{rubricT}{HTML}{166534}

\newtcolorbox{rubriccat}[2]{%
  enhanced, breakable,
  colback=#2!4, colframe=#2!75!black, colbacktitle=#2!85!black,
  coltitle=white,
  boxrule=0.5pt, arc=2pt,
  left=6pt, right=6pt, top=4pt, bottom=4pt,
  fonttitle=\bfseries,
  title={#1},
}

\newtcolorbox{rubriccatmuted}[2]{%
  enhanced, breakable,
  colback=#2!2, colframe=#2!55!white, colbacktitle=#2!70!white,
  coltitle=black!75,
  boxrule=0.4pt, arc=2pt,
  left=6pt, right=6pt, top=4pt, bottom=4pt,
  fonttitle=\bfseries,
  title={#1},
}

\NewEnviron{ack}{\BODY}

\title{Automated Benchmark Auditing for AI Agents and Large Language Models}
\author{%
  Junlin Wang$^{1}$, Federico Bianchi$^{2}$, Shang Zhu$^{2}$, Fan Nie$^{3}$, Yongchan Kwon$^{2}$, Bhuwan Dhingra$^{1}$, James Zou$^{2,3}$ \\
  $^{1}$Duke University, $^{2}$Together AI, $^{3}$Stanford University
}

\begin{document}

\maketitle

% !TEX root = ../neurips_2026.tex
\begin{abstract}
Modern AI benchmarks operate at a complexity that outpaces traditional verification methods. Tasks authored by domain experts often contain implicit assumptions, incomplete environment specifications, and brittle evaluation logic that human annotation cannot reliably catch. We introduce Auto Benchmark Audit (ABA), an agentic framework that systematically audits individual benchmark tasks, uncovering issues such as hidden environment dependencies, specification gaps, and limited grading logic. We run ABA on a collection of frontier LLM benchmarks and previous NeurIPS publications, totaling 168 benchmarks across nine domains. Across this corpus, ABA identifies critical issues including ambiguous task design, execution environment conflicts, and incorrect ground truths in over 25.7\% of the evaluated tasks. The precision of these automated audits is validated by expert review and independent third-party reports such as upstream PRs. Crucially, we demonstrate that these problematic tasks severely distort capability assessments for agents and LLMs: filtering out these tasks with issues shifts model rankings and increases average performance on SWE-bench Verified and Terminal-Bench 2 by 9.9\% and 9.6\%, respectively. We release the agentic tool and all task annotations to support the future development of frontier benchmarks.
\end{abstract}

\noindent \faIcon{globe} \textbf{Website:} \url{https://www.autobenchaudit.com/}

\noindent \faIcon{code} \textbf{Code:} \url{https://github.com/IsThatYou/auto-bench-audit}

% !TEX root = ../neurips_2026.tex
\section{Introduction}
\label{sec:introduction}

Scientific progress depends on the trustworthiness of its measurement instruments. In Artificial Intelligence, those instruments are benchmarks. If we cannot trust the methods we use to verify progress, our ability to define progress is undermined.

The AI community has long recognized the value of auditing benchmarks. Over the past decade, benchmark quality was maintained through sustained manual effort in different fields. Natural Language Inference is the canonical example: \citet{gururangan2018annotation} exposed systematic annotation artifacts introduced by crowd workers in the SNLI and MNLI benchmarks. The same pattern held in knowledge graph completion: \citet{akrami2020realistic} showed that data redundancy and test leakage in widely-used benchmarks caused accuracy to be largely overestimated. In each case, sustained manual effort was sufficient to identify and correct the problem, because the benchmarks themselves were within reach of generalist human effort. % \yc{we may want to polish this paragraph later. the point I was confused is this part: the opening sentence says `Benchmark quality was maintained through sustained manual effort,' so I expected some previous work on how the quality of benchmarks was maintained, but the canonical example with classification accuracy was less clear to me. Maybe we can directly say "Gururangan et al. considered systematic annotation by crowd workers....."}

Modern benchmarks like SWE-bench \citep{jimenez2024swebench} and Terminal-Bench \citep{merrill2026terminalbench} sit behind containerized environments, multi-stage evaluation harnesses, and grading logic that depends on runtime state. Tasks are contributed by domain experts who, paradoxically, are the least likely to make their assumptions explicit: they take critical context for granted, leaving setup requirements, environment specifications, and task constraints incomplete in ways invisible to them. Generalist annotation cannot fill this gap, and scaling up human effort does not solve a problem that is fundamentally one of expertise and coverage.

We introduce \textbf{Auto Benchmark Audit (ABA)}, an agentic framework for systematic issue detection in LLM benchmarks. Given a benchmark (i.e., its repository, dataset, and, when available, recorded agent trajectories), ABA provides structured findings under a single schema, each citing the file path from which it is drawn. A deterministic evidence collector resolves each benchmark's heterogeneous artifacts into a uniform manifest. The framework is benchmark-agnostic and operates across agentic, patch-based, and static-QA benchmarks within a unified protocol. We apply ABA to nine benchmark domains spanning over thirty thousand tasks, validate findings through upstream pull requests and independent human review, and release the full audit record as a public artifact. Beyond rediscovering known issues, ABA discovers more problematic benchmark tasks that include hidden environment conflicts, implicit specification gaps, or brittle evaluation logic, examples of which can be seen in Figure~\ref{fig:hook}.

Building a high-quality benchmark is hard, and the issues ABA identifies are not indictments of the teams who built them: they are the natural residue of the complexity of the current state of AI progress. We hope ABA can serve as a practical hardening tool: a way to catch what expert eyes miss, and to treat benchmarks as the living artifacts they need to be. In summary, our contributions are:

\begin{enumerate}[leftmargin=*,topsep=2pt,itemsep=2pt]
\item \textbf{An agentic auditing framework.} ABA combines static and trajectory auditing modes (analyzing benchmark artifacts directly vs. replaying recorded agent runs) with task-level audit scopes into a single benchmark-agnostic pipeline. The framework adapts to arbitrary evaluation harnesses and can suggest targeted fixes alongside each finding.
\item \textbf{Large-scale benchmark auditing and artifact release.} We apply ABA across 168 benchmarks across nine domains and over thirty thousand tasks, releasing all audit findings, per-task evidence, and rubric annotations as a public artifact.
\item \textbf{Empirical validation of ABA findings.} We confirm identified issues through upstream pull requests and meticulous manual reviews performed by domain experts. The framework presents task issues the even original authors had not previously identified, such as hidden environment dependencies, implicit specification gaps, and brittle grading logic.
\end{enumerate}

\newcommand{\hookhl}[2]{%
  {\setlength{\fboxsep}{1pt}\colorbox{#1!28}{\strut #2}}%
}
\newcommand{\hooksection}[3]{%
  % #1 color, #2 label, #3 content
  \begin{tcolorbox}[
    enhanced, width=\linewidth,
    colback=#1!50, opacityback=0.18,
    frame hidden, boxrule=0pt, arc=1.5pt,
    boxsep=0pt, left=3pt, right=3pt, top=1pt, bottom=1pt,
    fontupper=\scriptsize,
  ]
    \textcolor{#1!70!black}{\textbf{#2}} #3
  \end{tcolorbox}%
}
\newcommand{\hookcard}[6]{%
  % #1 color, #2 category, #3 benchmark, #4 prompt, #5 eval, #6 issue
  \begin{tcolorbox}[
    enhanced, width=\linewidth,
    equal height group=hookrow,
    colback=white, colframe=#1!70!black,
    colbacktitle=#1!80!black, coltitle=white,
    boxrule=0.5pt, arc=2pt,
    boxsep=0pt, left=3pt, right=3pt, top=1pt, bottom=1pt,
    toptitle=1pt, bottomtitle=1pt,
    fonttitle=\bfseries\footnotesize,
    title={\textsc{#2}\hfill\normalfont\scriptsize\textsf{#3}}
  ]
    \hooksection{#1}{Prompt.}{#4}\vspace{-7pt}
    \hooksection{#1}{Evaluation.}{#5}\vspace{-7pt}
    \hooksection{#1}{Issue.}{#6}
  \end{tcolorbox}%
}

\begin{figure}[t]
  \centering
  \begin{minipage}[t]{0.325\linewidth}
    \hookcard{rubricA}{Instruction}{FinanceAgent}
      {``Compare KO's FY24 dividend payout ratio to its \hookhl{rubricA}{competitors}\,\dots\, rank highest-to-lowest.''}
      {Per-ticker exact-value checks against one fixed peer set
      (KDP, KO, PEP, KHC, SJM).}
      {``Competitors'' is never defined; the rubric pins one peer set,
      omitting MNST (KO's canonical direct competitor) and including
      two non-beverage firms.}
  \end{minipage}\hfill
  \begin{minipage}[t]{0.325\linewidth}
    \hookcard{rubricE}{Environment}{Terminal-Bench~2}
      {``\hookhl{rubricE}{Download this video} of zork [\textsf{youtube.com/\dots}]\,\dots\, write all the moves to \texttt{/app/solution.txt}.''}
      {Levenshtein similarity $\geq 90\%$ to a hardcoded keystroke transcript.}
      {Bare \texttt{ubuntu:24.04}, no cached video; YouTube returns HTTP~403
      to cloud IPs. 4 of 5 runs fail before \texttt{solution.txt} is written.}
  \end{minipage}\hfill
  \begin{minipage}[t]{0.325\linewidth}
    \hookcard{rubricT}{Evaluation}{OSWorld-V}
      {``Take down meta info of HF papers on 1 Mar 2024 in the \texttt{.docx} file\,\dots\, \hookhl{rubricT}{conform to the format} and complete others.''}
      {Exact-string comparison against reference docx files (whitespace/case normalized).}
      {The reference inconsistently mixes \texttt{/pdf/} vs. \texttt{/abs/} URL formats; agents faithfully following the stated format fail the exact-match evaluation.}
  \end{minipage}
  \caption{Examples of task-level issues identified by ABA. \textbf{Instruction}: the prompt
  underdetermines the answer the rubric demands. \textbf{Environment}:
  the container lacks the resources the prompt assumes. \textbf{Evaluation}: the grading harness rejects solutions that faithfully
  follow the prompt. }
  \label{fig:hook}
\end{figure}

% !TEX root = ../neurips_2026.tex
\section{Auto Benchmark Audit Overview}
\label{sec:framework}

\begin{figure}[t]
  \centering
  \hspace*{0.03\linewidth}\includegraphics[width=0.97\linewidth]{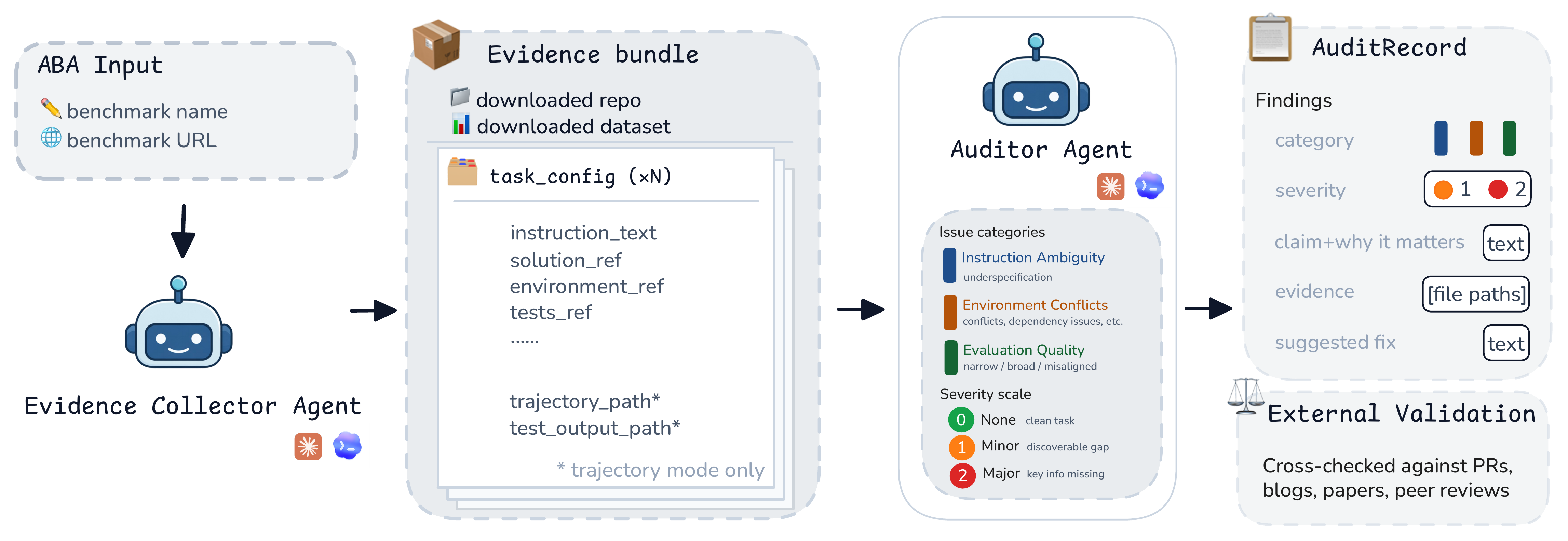}
  \caption{The Auto Benchmark Audit pipeline. Given benchmark name and URL, an Evidence Collector Agent downloads necessary files and resolves each
  benchmark's heterogeneous artifacts into a uniform manifest of
  file paths. Then an Auditor Agent audits based on issue categories and severity scale. Then findings include specific evidences, reasoning and suggested fix.}
  \label{fig:framework}
\end{figure}

Auto Benchmark Audit (ABA) takes a benchmark (its name and URLs to the dataset or repository) and outputs structured records of the benchmark issues.
Figure~\ref{fig:framework} provides an overview of the pipeline, and the remainder of this section describes each component in detail.

% \paragraph{Anatomy of a Benchmark Task }\label{sec:framework:anatomy} 
% Every benchmark we consider is decomposed into three components:
% (1) \text{Instruction}:
% The natural-language prompt the LLM system receives: In static QA it may be a single question; in agentic tasks it is a multi-paragraph specification describing the goal, the tools and file descriptions.
% (2) \text{Environment}:
% The artifacts and execution surface the agent operates over:
% \emph{static} (text, files, and images the prompt references),
% \emph{executable} (interpreters, containers, and repositories the
% agent runs code against), and \emph{interactive} (browsers, tool
% servers, and simulators).
% (3) \text{Evaluation}:
% The mechanism that turns an agent or LLM's output into a pass/fail
% signal: \emph{deterministic} graders (exact match, programmatic
% checks, and scalar metrics, all of which require ground-truth answers to compare against), \emph{execution-based} graders (running
% the agent's code, unit tests, or state assertions), or
% \emph{subjective} graders (LLM-as-judge, human review).

\paragraph{Evidence Collector Agent}\label{sec:framework:protocol}
The first challenge in developing an automated benchmark auditing pipeline is how to handle highly heterogeneous configurations of existing benchmark tasks. Task representations vary drastically across benchmarks: they may appear as a directory of files in Terminal-Bench~2, as a \texttt{task\_patch} embedded within a JSON in SWE-bench, or as simple \texttt{(instruction, gold\_answer)} pairs for static QA datasets like HLE \citep{phan2025lastexam} and GPQA \citep{rein2024gpqa}. Furthermore, certain tasks require complex environmental dependencies, such as accompanying container images.

To handle this heterogeneity, ABA uses a deterministic manifest schema with an evidence collector agent shown in \Cref{fig:framework}. The agent first downloads required files like GitHub repositories or Huggingface datasets. Then for each task, this agent maps heterogeneous benchmark artifacts into a unified representation. Lightweight data, such as \texttt{instruction\_text}, is inlined directly, while heavyweight assets and execution traces are isolated via filesystem references (e.g., \texttt{solution\_ref}, \texttt{environment\_ref}, and \texttt{trajectory\_path}). By abstracting away the underlying data structures, onboarding a new benchmark is reduced to providing a minimalist configuration containing the benchmark's name and repository URL. The collector automatically adapts this source into the standardized ABA schema for agentic inspection, allowing us to scale the evaluation suite to 168 distinct benchmarks without any per-benchmark modifications to prompts, rubrics, or output schemas. The evidence collector agent prompt can be found in \Cref{app:collector_prompts}.

\paragraph{Auditor Agent}
\label{sec:framework:auditing_protocol}
After evidence collection, an auditor agent is initialized with the task configuration and file paths pointing to the relevant repositories and datasets. Using shell tools, the auditor inspects these artifacts and analyzes the task guided by a standardized rubric (full prompt templates are detailed in Appendix~\ref{app:prompts}). This auditing procedure remains uniform across all task types, from static QA to multi-step agent environments. 

The protocol runs in one of two modes that differ in what evidence is gathered:
(1) \textbf{Static mode}, where the auditor reads only the task definition --- the instruction the agent would receive, the tests against which its solution would be graded, the evaluation configuration, etc., with no execution observed.
(2) \textbf{Trajectory mode}, where the auditor additionally reads recorded agent traces (\texttt{trajectory\_path}) and the test output result (\texttt{test\_output\_path}), etc. enabling the auditor to identify runtime issues that static inspection cannot capture.
In Section \ref{sec:ablations}, we show that the trajectory mode finds more issues than the static mode by leveraging intermediate execution traces and runtime behavior.

\begin{figure}[h]
  \centering
  \begin{tcolorbox}[
    enhanced, colback=white, colframe=black!55,
    title={\textbf{Task-level issue categories}\hfill
           \normalfont\footnotesize 
           },
    fonttitle=\bfseries, coltitle=white, colbacktitle=black!75,
    boxrule=0.4pt, arc=2pt,
    left=6pt, right=6pt, top=5pt, bottom=5pt,
    fontupper=\footnotesize
  ]
  \begin{tabularx}{\linewidth}{@{}>{\centering\arraybackslash}X
                                 >{\centering\arraybackslash}X
                                 >{\centering\arraybackslash}X@{}}
    \textcolor{rubricA}{\textbf{ Instruction}} &
    \textcolor{rubricE}{\textbf{Environment}} &
    \textcolor{rubricT}{\textbf{Evaluation}} \\
  \end{tabularx}
  \vspace{-5pt}
  \begin{tcbraster}[
    raster columns=3, raster equal height=rows,
    raster column skip=4pt, raster row skip=4pt,
    enhanced, boxrule=0.3pt, arc=1.5pt,
    boxsep=0pt, left=4pt, right=4pt, top=3pt, bottom=3pt,
    fontupper=\footnotesize
  ]
    \begin{tcolorbox}[colback=rubricA!4, colframe=rubricA!50]
    The prompt is missing critical information, is ambiguous, or is
    misleading in a way that drives failures. The reader
    cannot determine what passing-the-test means from the prompt alone.
    \end{tcolorbox}
    \begin{tcolorbox}[colback=rubricE!4, colframe=rubricE!50]
    The runtime environment --- container image, resources, installed
    tools, filesystem --- conflicts with what the task instructions
    assume or require, blocking the approach the prompt instructs.
    \end{tcolorbox}
    \begin{tcolorbox}[colback=rubricT!4, colframe=rubricT!50]
    The test suite does not fairly and completely evaluate correct
    solutions: too narrow (rejects valid alternatives), too broad
    (accepts trivially wrong outputs), or misaligned with the stated
    objective. Or simply, the ground-truth answer is not correct.
    \end{tcolorbox}
    \begin{tcolorbox}[colback=rubricA!8, colframe=rubricA!50]
    \textcolor{rubricA}{\textit{Example.}} The prompt says ``sort the
    records'' but the test accepts only descending order by a secondary
    key the prompt never names; two reasonable readers produce two
    different solutions, only one of which passes.
    \end{tcolorbox}
    \begin{tcolorbox}[colback=rubricE!8, colframe=rubricE!50]
    \textcolor{rubricE}{\textit{Example.}} The prompt instructs the
    agent to inspect git history with \texttt{git log}, but
    \texttt{git} is not installed in the container and the repo ships
    as a flat snapshot, so the prescribed approach is impossible.
    \end{tcolorbox}
    \begin{tcolorbox}[colback=rubricT!8, colframe=rubricT!50]
    \textcolor{rubricT}{\textit{Example.}} The test checks the
    program's stdout against an exact-string fixture including a
    trailing newline; a functionally correct solution that prints the
    same content without the trailing newline is marked failing.
    \end{tcolorbox}
  \end{tcbraster}
  \end{tcolorbox}
  \caption{Task-level issue categories. The issues can be divided into three independent axes: Instruction, Environment, and Evaluation.}
  \label{fig:task_defect_categories}
\end{figure}

\paragraph{Audit Findings}
\label{sec:framework:severity_calibration}
As shown in Figure~\ref{fig:framework}, each finding carries the following five fields.
(1) \text{category}: every identified issue is categorized along three independent axes, Instruction, Environment, and Evaluation. Detailed category definitions are shown in \Cref{fig:task_defect_categories}.
(2) \text{severity}: a 3-class score [0, 1, 2], where 2 denotes a major issue that renders the task unsolvable (e.g., conflicting instruction constraints), 1 denotes a minor issue that can be resolved with sufficient domain expertise, and 0 indicates no issue.
(3) \text{evidence}: the filesystem paths or artifacts supporting the claim;
(4) \text{claim \& why it matters}: a natural-language description of the identified issue plus the downstream failure mode induced by the issue in the evaluation.
(5) \text{suggested fix}: a concrete change to the prompt, environment, or tests, in order to resolve the issue.
% Note that although our focus is on task-level benchmark audits, the ABA framework can be extended to whole-benchmark-level audits as well.

Finally, ABA findings are validated through cross-referencing with external evidence sources, including upstream pull requests, published papers and blogs, and expert manual review.

% \input{sections/experiment_setup}

% !TEX root = ../neurips_2026.tex
\section{Auditing Benchmark Quality}
\label{sec:audit_results}

\subsection{Experiment Setup}
\label{sec:audit:experiment_setup}

To test the effectiveness of ABA, we draw benchmarks from two complementary sources: (1) Frontier Benchmarks\label{sec:audit:experiment_setup:benchmarks}
(Appendix~\ref{app:portfolio:frontier}, Fig.~\ref{fig:portfolio:funnel}): we take commonly used benchmarks from five recent frontier LLM releases.\footnote{Anthropic Opus~4.7~\citep{anthropic2026opus47}, OpenAI
GPT-5.4~\citep{openai2026gpt54}, Zhipu GLM-5.1~\citep{glm5team2026glm5}, Moonshot
Kimi~K2.6~\citep{moonshot2026kimik26}, MiniMax M2.7~\citep{minimax2026m27}. }
(2)  NeurIPS 2025 accepted benchmarks: we collect and filter NeurIPS 2025 Datasets \& Benchmarks Track accepted papers, more details of which can be seen in Appendix~\ref{app:portfolio:neurips2025}. In total, 168 benchmarks are collected from these two sources, where we maximize task coverage as much as possible allowed by feasible budget constraints. The details are provided in \Cref{tab:appendix:per_benchmark_overview}.

\paragraph{Benchmark filtering}
We select benchmarks based on two primary criteria. First, the system under consideration must be general-purpose LLMs or LLM agents. Therefore, we exclude specialized domains or modalities, such as audio, embodied AI, and remote sensing. Second, each task needs to carry a verifiable ground truth (e.g. test suite, gold answer, or deterministic grader) which the audit rubric can operate on. Therefore, we exclude those with inherently subjective evaluation. We also exclude meta-evaluation benchmarks which assess evaluation methodologies. More details can be found in Appendix~\ref{app:portfolio:neurips2025}.

\paragraph{Agent configuration}
\label{sec:audit:experiment_setup:configuration}
We run ABA through the Claude Code CLI with its default settings: Opus 4.7, default tool permissions (file read/write, bash, and search), and the default system prompt and agent loop. We fix the version at v2.1.96.

\subsection{ABA Findings}
\label{sec:audit:portfolio}

% !TEX root = ../neurips_2026.tex
% Full-coverage domain-level severity rollup -- companion to
% \texttt{tables/coverage\_domains.tex} that extends the row counts to
% every benchmark in
% \texttt{benchmarks/multi-domain/benchmark\_categorization.json}
% (i.e.\ the curated paper portfolio plus the NeurIPS 2026 D\&B sweep
% stored under \texttt{data/bench\_audit/multi\_domain\_all/neurips/}).
% Each row pools all audited tasks within the domain across constituent
% benchmarks for the **static** audit only; trajectory-only runs are
% excluded so the per-domain task counts and severity shares reflect a
% single audit mode. Severity bins are the v3 task-level rubric --
% Clean / Minor / Major (no Critical bucket; cf.
% \S\ref{sec:framework:rubric}); the small number of tasks recorded
% at sev.\,3 in v3 runs are folded into Major. Numbers are sourced
% from \texttt{visualizer/public/data/index.json}
% (\texttt{overviewStats.static.taskSeverityDist}) and benchmark
% membership/domain labels from
% \texttt{benchmarks/multi-domain/benchmark\_categorization.json}
% and regenerated by \texttt{draft\_figures/scripts/coverage\_domains\_full.py}.

\begin{table}[t]
  \centering
  \small
  \caption{Overview of ABA results by domain (static-mode audit only).
  Each row reports the number of benchmarks and tasks. We report both the number of major issues and affected tasks as one task may contain multiple issues. Additionally, \textit{Severity Share} column categorizes tasks by their highest-severity issues: the \textit{Major} column indicates the percentage of tasks with at least one major finding, while the \textit{Minor} column shows the percentage of tasks where the most severe finding is minor.}
  \label{tab:coverage:domains:full}
  \begin{tabular}{lrrrrrr}
    \toprule
    & & &
      \multicolumn{2}{c}{\textbf{Major Issues}} &
      \multicolumn{2}{c}{\textbf{Severity Share}} \\
    \cmidrule(lr){4-5} \cmidrule(lr){6-7}
    \textbf{Domain} &
      \textbf{Benchmarks} &
      \textbf{Tasks} &
      \textbf{Issues} &
      \textbf{Tasks} &
      \textbf{Major} &
      \textbf{Minor} \\
    \midrule
    Science        &  15 & 3{,}233 & 1{,}120 &     831 & 25.7\,\% & 16.8\,\% \\
    Multimodal     &  37 & 8{,}330 & 3{,}264 & 2{,}054 & 24.7\,\% & 13.3\,\% \\
    Professional      &   8 & 1{,}160 &     602 &     445 & 38.4\,\% & 24.4\,\% \\
    Agentic / Tool Use  &  23 & 3{,}337 &     898 &     686 & 20.6\,\% & 20.3\,\% \\
    Coding           &  24 & 5{,}041 &     877 &     713 & 14.1\,\% & 14.2\,\% \\
    Medical                &  17 & 4{,}014 & 3{,}314 & 1{,}664 & 41.5\,\% & 13.3\,\% \\
    Math           &  14 & 3{,}060 &     490 &     403 & 13.2\,\% &  8.1\,\% \\
    Retrieval / RAG          &   8 & 2{,}210 &     495 &     377 & 17.1\,\% & 21.1\,\% \\
    Safety / Alignment                &  22 & 3{,}900 & 2{,}964 & 1{,}646 & 42.2\,\% & 15.7\,\% \\
    \midrule
    \textbf{Total}        & 168 & 34{,}285 & 14{,}024 & 8{,}819 & 25.7\,\% & 15.1\,\% \\
    \bottomrule
  \end{tabular}
\end{table}

According to Table~\ref{tab:coverage:domains:full}, our static-mode audit of 168 benchmarks and 34,285 tasks reveals that 25.7\,\% of them carry major issues (severity 2) under the task-level rubric while 15.1\% contain minor issues (severity 1) but no major issues, so less than 60\% of the tasks remain clean (severity 0).

One task may have multiple issues, as reflected in Table~\ref{tab:coverage:domains:full}, where the \textit{Issues} column tallies the total number of major issues, while the \textit{Tasks} column reports the number of unique tasks affected by at least one major issue.

Across benchmarks, the gap between the highest and lowest major severity percentage spans more than a factor of three. The Math domain sees the lowest major severity rate
(13.2\% Major, 78.7\% Clean), where tasks usually specify
numeric answers and deterministic graders, while no environments or test scripts
are required. The Coding domain is the second lowest with only 14.1\% Major. In contrast, the Safety / Alignment (42.2\% Major),
Medical (41.5\% Major), and Professional (38.4\% Major) domains display the highest major severity shares. Note that they account for 42.6\% of tasks with major issues despite only 26.5\% of all audited tasks. Qualitatively, the major issue shares associate strongly with the complexity of tasks' environments and evaluation setups. Domains requiring intricate environments or nuanced grading protocols are more prone to unstated assumptions.

\subsection{Issue Patterns across Domains}
\label{sec:audit:signatures}

\begin{figure}[t]
  \centering
  \includegraphics[width=\linewidth]{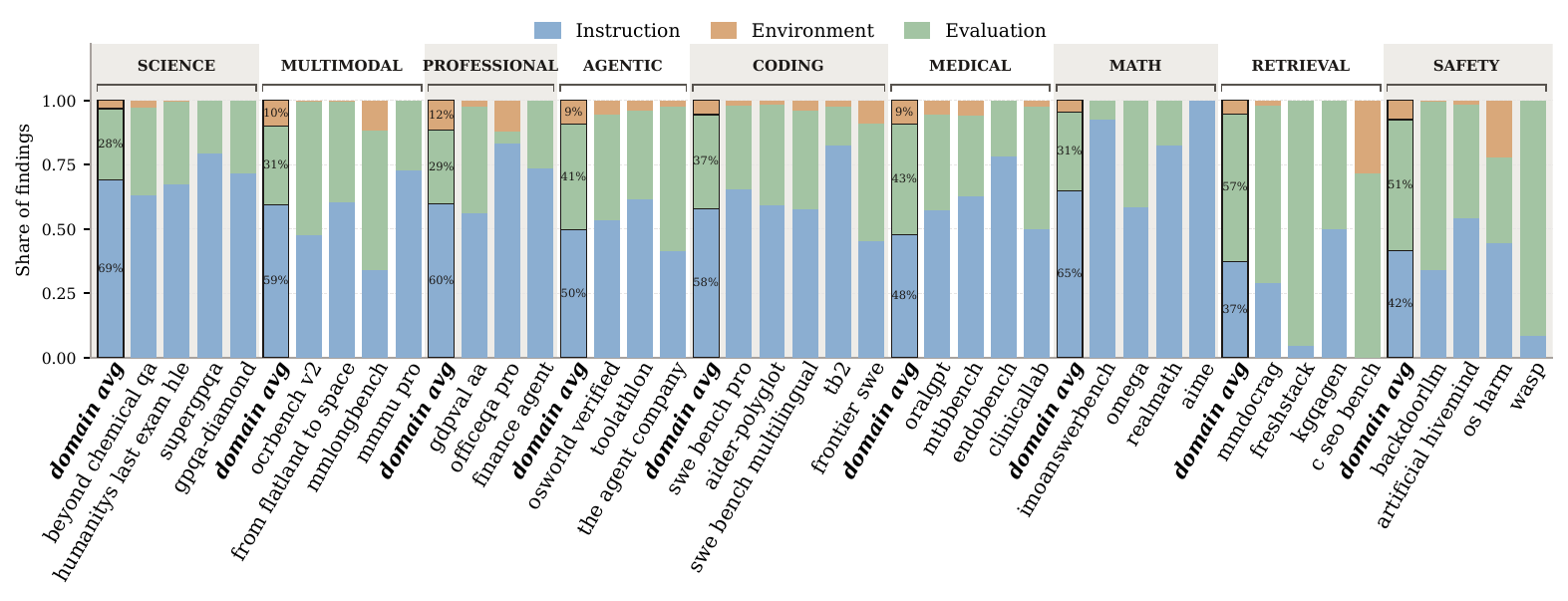}
  \caption{Issue categories 
  per audited benchmark, grouped by domain, for all non-clean static-audit findings. Each finding is one of Instruction, Environment, or Evaluation issues. We present the domain average and a subset of covered benchmarks in each domain.}
  \label{fig:audit:aet_distribution}
\end{figure}
This subsection breaks down the major issue rate by issue types (instruction, environment, evaluation) and we discuss three main patterns across domains. Further, \Cref{fig:audit:aet_distribution} illustrates the percentage decomposition and displays a representative subset of benchmarks from each domain.

\paragraph{Domains with deterministic evaluations are instruction-issue dominant.}
When the evaluation method is based on deterministic checkers (e.g., a numeric answer, an exact-match string, or an option in multiple-choice questions), instruction ambiguity emerges as the primary failure source. The Math domain is a clear illustration, as 65\% major issues are caused by instructions, with certain benchmarks exceeding 90\% (e.g. IMOAnswerBench \citep{luong-etal-2025-towards}). The Science domain follows a similar pattern, with instruction-issues accounting for 69\% of total. In these two domains, over a quarter of the issues arise from evaluation setups such as incorrect gold references, while they exhibit the lowest rate of environment-issues, which is close to 0.

\paragraph{The safety and retrieval domains are evaluation-issue dominant.}
The Safety and Retrieval domains are dominated by evaluation-based issues, which represent 51\% and 57\% of total, respectively. This can be attributed to the fact that tasks in these domains frequently require multi-faceted answers, so their evaluation suites necessitate complex grading mechanisms, such as LLM-as-a-judge and test rubrics. This introduces issues such as unreliable LLM judges and inadequate test coverage.

\paragraph{Complex runtimes introduce environment vulnerabilities.} Environment-related errors associate with the complexity of execution requirements. While static-QA tasks naturally exhibit near-zero environment errors, domains requiring complex runtimes face severe vulnerabilities. Consequently, the Agentic, Professional, and Multimodal domains exhibit higher percentages of environment-based issues across all categories.

% !TEX root = ../neurips_2026.tex
\section{Validating ABA Results}
\label{sec:audit:validation}

Ground truth is extremely scarce in this domain, and it is a non-trivial task to find a clean test set for this type of problem. The experts who could produce one are typically the same people who authored the benchmarks, and the issues we care about are precisely those they might have overlooked. We therefore organize our validation around a hierarchy of evidence. At the top is what we call \emph{fixes in the wild}: cases where practitioners independently decided an issue was serious enough to act on by issuing a fix PR or retiring a benchmark from evaluation entirely. These cases carry the strongest epistemic weight because the stakes are real and the motivation is not evaluative. Below them are two constructed ground-truth sources: the BenchGuard gold-issue sets for BixBench and ScienceAgentBench (Section \ref{sec:audit:validation:benchguard}), curated carefully but for auditing purposes, and an internal human-rater study on major issues found by ABA (Section \ref{sec:audit:validation:manual}).

\subsection{Fixes in the Wild}
\label{sec:audit:validation:wild}

Our strongest validation comes from cases where practitioners independently acted on benchmark issues without sight of our findings. We draw on two instances.

\paragraph{Maintainer fix PR (Terminal-Bench~2).} The Terminal-Bench~2 maintainers proposed a fix PR\footnote{\url{https://github.com/harbor-framework/terminal-bench-2/pull/53}} targeting 22 tasks. Four task fixes are simple due diligence fixes about task resources or oracle solutions (\Cref{app:tb2_pr53}). Of the remaining 18 we decompose 21 individual issues. Our trajectory audit independently recovered 14 of these 21 issues under strict matching and 17 under partial matching. This yields a strict recall of 66.7\% (81.0\% partial) against the maintainers' ground truth—a set of fixes they developed entirely independent of our findings (top row of Table~\ref{tab:audit:benchguard}). Precision measured only against PR\,\#53 is 58.3\,\%\,/\,70.8\,\%, but this is a lower bound because the PR is not exhaustive. ABA also flags 11 tasks outside the PR; on the seven Major tasks we review by hand, its 11 findings achieve 72.7\,\% strict and 81.8\,\% aligned-or-partial precision. The PR overlap therefore validates two contributions at once: ABA recovers many defects that maintainers independently fixed, and it surfaces additional real defects beyond the scope of that human review. Per-task lists for both sets are in Appendix~\ref{app:tb2_pr53}.

\paragraph{Research blog (SWE-bench Verified).} A recent OpenAI post explaining their decision to retire SWE-bench Verified from internal evaluation\footnote{\url{https://openai.com/index/why-we-no-longer-evaluate-swe-bench-verified/}} singles out two tasks -- \texttt{pylint-dev\_\_pylint-4551} and \texttt{sympy\_\_sympy-18199} -- as concrete examples of unreliable grading. Both present as major issues in our audit, flagged independently and without reading the blog post. This instance differs from the PR case in a consequential way: the action taken was not a patch but a retirement, meaning the practitioners judged the issues severe enough to abandon one of the field's most-reported coding benchmarks entirely. This study underscores the practical value of ABA in identifying critical benchmark issues. For both cases, our auditor never queried Git histories, never fetched the related PR, and never executed web searches.

\subsection{A Recent Study on Scientific Benchmarks (BenchGuard)}
\label{sec:audit:validation:benchguard}

% !TEX root = ../neurips_2026.tex
% Unified external validation table. Sources:
%   - Terminal-Bench 2 maintainer PR #53
%     (https://github.com/harbor-framework/terminal-bench-2/pull/53):
%     22 PR-flagged tasks, 18 in-scope under task-level rubric;
%     ABA recovers 14/18 (77.8%) and contributes 11 additional
%     ABA-only findings. Precision against the PR is omitted because
%     those 11 ABA-only findings may be real issues (manually
%     confirmed in \S\ref{sec:audit:validation:manual}).
%   - BenchGuard reports for BixBench / SAB:
%       bench_audit/benchmarks/benchguard/output/reports/bixbench_eval.md
%       bench_audit/benchmarks/benchguard/output/reports/sab_eval.md
%   - Opus 4.6 definition-only baselines from Tu et al. 2026
%     Tables 2 (SAB) and 3 (BIXBench Verified-50).

\begin{table}[t]
  \centering
  \small
  \setlength{\tabcolsep}{4pt}
  \caption{External validation. Recall and precision (\%) under
  strict (s) and partial (p) match
  criteria for the three gold-issue sets from Terminal-Bench 2, BixBench, ScienceAgentBench. Redund.\
  is the average number of aligned findings emitted per unique gold issue caught: a higher value indicates the auditor reports multiple findings against the same issue. Note that Claude Opus 4.6 is used here to match the reference settings of BenchGuard.}
  \label{tab:audit:benchguard}
  \begin{tabular}{lllrrccc}
    \toprule
    \textbf{Benchmark} &
      \textbf{Auditor} &
      \textbf{Model} &
      \textbf{Gold} &
      \textbf{Find.} &
      \textbf{Recall\ (s\,/\,p)} &
      \textbf{Precision \ (s\,/\,p)} &
      \textbf{Redund.} \\
    \midrule
    Terminal-Bench~2
      & ABA (ours)                         & Opus~4.6 & 21 & 24 & 66.7 / 81.0 & 58.3 / 70.8 & 1.0$\times$ \\
    \midrule
    \multirow{2}{*}{BixBench}
      & ABA (ours)                         & Opus~4.6 & 24 & 28 & \textbf{62.5 / 79.2} & \textbf{50.0 / 71.4} & \textbf{1.05$\times$} \\
      & BenchGuard & Opus~4.6 & 24 & 31 & 54.2 / 79.2 & 38.7 / 67.7 & 1.10$\times$ \\
    \midrule
    \multirow{2}{*}{ScienceAgentBench}
      & ABA (ours)                         & Opus~4.6 & 12 & 27 & \textbf{83.3 / 91.7} & 40.7 / 51.9 & \textbf{1.40$\times$} \\
      & BenchGuard & Opus~4.6 & 12 & 31 & \textbf{83.3 / 91.7} & --\, /\, \textbf{58.1} & 1.64$\times$ \\
    \bottomrule
  \end{tabular}
\end{table}

For a complementary check we turn to BenchGuard \citep{tu2026benchguard}, which curates gold-issue sets for two execution-based benchmarks:  ScienceAgentBench \citep{chen2025scienceagentbench} and the
BixBench Verified-50 subset \citep{mitchener2025bixbench0}. We run ABA on the same two task sets and report alignment against the same gold issues using the same alignment judge, alongside the strongest single-model baseline BenchGuard reports (Claude Opus~4.6 in definition-only mode). We run ABA with Claude Opus~4.6 to ensure fair comparison to BenchGuard.

On ScienceAgentBench, ABA recovers 10/12 gold issues strictly and 11/12 with
partial credit, matching the Opus~4.6 baseline at both bars. 
% The single missed issue (\texttt{67}) concerns a column-naming convention on an otherwise fully-specified output; it sits at the boundary between ambiguity and stylistic preference. 
 On BixBench, ABA's strict recall improves over the BenchGuard single-model
baseline (62.5\,\% vs.\ 54.2\,\%) at matched partial recall
(79.2\,\%), and its precision is higher under both match criteria
(50.0\,\%\,/\,71.4\,\% vs.\ 38.7\,\%\,/\,67.7\,\%). The redundancy
column tells the same story from a different angle: BenchGuard tends
to emit multiple findings against the same gold issue (1.10$\times$
on BixBench, 1.64$\times$ on ScienceAgentBench), while ABA reduces redundancy from 1.10$\times$ to
1.05$\times$ on BixBench and from 1.64$\times$ to 1.40$\times$ on ScienceAgentBench.

\subsection{Manual Confirmation of Major Issues}
\label{sec:audit:validation:manual}

% !TEX root = ../neurips_2026.tex
\setlength{\intextsep}{0pt}
\begin{wraptable}{r}{0.5\textwidth}
  % \vspace{}
  \centering
  \small
  \setlength{\tabcolsep}{4pt}
  \caption{Manual confirmation of sampled issues (static mode) from 12
  benchmarks and 25 major issues (trajectory mode) sampled from SWE-bench Verified. Precision under strict~(s)
  and partial~(p) match.}
  \label{tab:per_task:manual_confirmation}
  \begin{tabular}{lccc}
    \toprule
    \textbf{Tier} & \textbf{Find.} &
      \textbf{Prec.\ (s)} & \textbf{Prec.\ (p)} \\
    \midrule
    Major & 56  & 73\%  & 91\% \\
    Minor & 30  & 63\% & 83\% \\
    \midrule
    SWE-bench Verified  & 25 & 92\% & 96\% \\
    \bottomrule
  \end{tabular}
  \vspace{0.1em}
\end{wraptable}

To further validate the audit at the task level, we manually review a sample of 56 major issues
and 30 minor issues in static mode. They span across 12 frontier benchmarks, where current capability claims are
concentrated and a spurious finding would be most consequential (list of 12 benchmarks at \Cref{app:validation:manual_benchmarks}). We also sample 25 major issues in trajectory mode from SWE-Bench Verified, as shown in
Table~\ref{tab:per_task:manual_confirmation}.

% !TEX root = ../neurips_2026.tex
\section{Trajectory Audit Analysis}
\label{sec:ablations}

Previous sections focused primarily on static-mode audits. We now examine the trajectory audit mode, which has access to richer information (e.g., runtime behavior and execution traces) and can therefore capture issues that are inaccessible from the static mode alone. This section compares the two audit modes and studies how trajectory-identified issues affect benchmark leaderboard rankings. 

\begin{figure}[t]
  \centering
  \includegraphics[width=\linewidth]{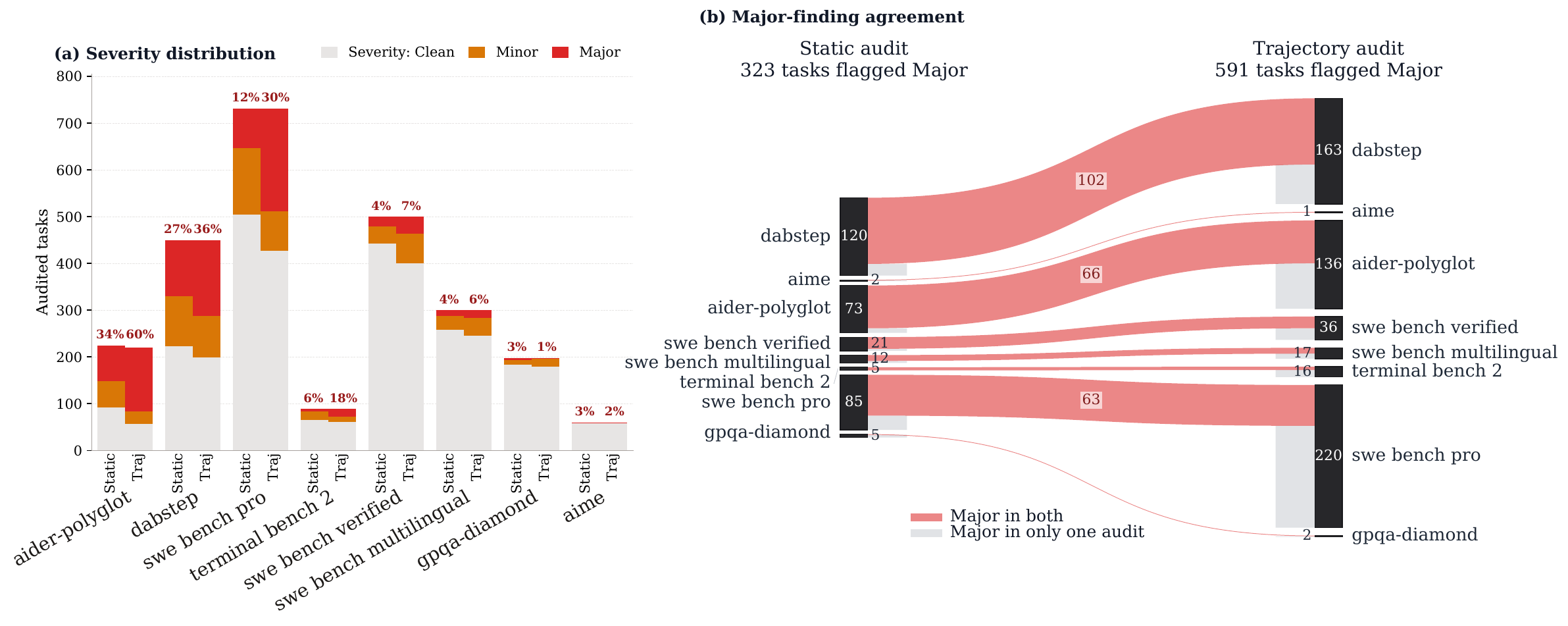}
  \caption{Static-vs-trajectory audit comparison on the benchmarks
  audited in both modes. (a) Severity distribution per benchmark at task-level. Each task is counted as a major task if it has at least one major issue. It is a minor task if most severe finding is minor. (b) Agreement between the two modes on shared tasks.} % summarised by \(\overline{p}_{\max}\) (share of trajectory flagged tasks also flagged by static at max severity)
  \label{fig:ablation:severity_and_sankey}
\end{figure}

\paragraph{Trajectory mode identifies more major issues.}
Figure~\ref{fig:ablation:severity_and_sankey} compares the static and trajectory audit modes across eight benchmarks. Trajectory mode consistently finds more major issues than the static mode on nearly every benchmark, along with a slightly larger overall non-clean share. Specifically, the trajectory mode flags 8.5\% more major tasks (tasks that have at least one major issue) and 2.4\% less minor tasks than the static mode on average across the 8 benchmarks. That is +6.1\% more tasks flagged overall. Most of the gap comes from the major issues, suggesting that the additional runtime evidence available to the trajectory mode exposes more consequential issues (e.g., silent failure modes and environmental drift) that are inaccessible to the static mode alone. 

\paragraph{Static and trajectory modes yield the identical audit results for many tasks.}
Figure~\ref{fig:ablation:severity_and_sankey} also illustrates the degree of overlap between the two audit modes. We find the max-severity agreement\footnote{We define max-severity agreement as the fraction of trajectory-identified major issues that are also flagged as major by the static mode.} ranges from 29\% to 63\%, with the average of 53\%. Here, we exclude {\tt aime} due to its small sample size. The disagreement largely results from the available information difference: runtime contamination and environment drift emerge only in the trajectory mode, while metadata ambiguity and test-suite leakage are primarily visible from the static mode. Despite these differences, the substantial overlap suggests that static audits already capture many of the same high-severity issues identified by the trajectory mode, supporting the robustness of the static mode. 

\paragraph{Audit results affect public leaderboard rankings.}
We next investigate how benchmark leaderboard rankings shift when audit-flagged issues are removed. We consider the five benchmarks, namely {\tt dabstep}, {\tt swe\_bench\_pro}, {\tt swe\_bench\_verified},
{\tt tb2}, {\tt swe\_bench\_multilingual}, and evaluate the mean per-model ranking change. The overall mean lifts for trajectory audit is 10.5 percent when filtering both major and minor issues. In particular, mean lifts reach roughly 12.8 percentage points on Dabstep \citep{egg2025dabstep0} when both major and minor issues are excluded, with several-point increases remaining even under the conservative major-only filter (see Figure~\ref{fig:ablation:leaderboard_mean_delta}). We also include an actual model ranking shift example in \Cref{app:leaderboards:shifts}. These shifts indicate that audit findings can significantly affect leaderboard rankings. We defer to Appendix~\ref{app:leaderboards} the full per-benchmark figure and a hypothesis test that the trajectory auditor's additional findings are not driven by agent failure.

\section{Related Work}
\label{sec:related_work}

Our work intersects three lines of research: systematic critique of
benchmark validity, benchmarks for LLM and agent capabilities, and
LLM-based automated evaluation.

\subsection{Benchmark Quality, Validity, and Critique}

A growing body of work audits individual benchmarks to expose
issues that distort reported scores. \citet{aleithan2024swebenchplus}
report 32.7\% solution leakage and order-of-magnitude inflation
from weak tests in SWE-bench, \citet{yu2025utboost} find
5.2/7.7\% of SWE-bench Verified/Lite tasks have insufficient tests, and \citet{garg2025savingswebench} show
benchmark mutation reveals over-fitting that inflates measured
capability. Beyond software engineering,
\citet{lange2025aicuda} identify grading flaws in
KernelBench \citep{DBLP:conf/icml/Ouyang0AZHRM25} that let agents bypass correctness checks,
and \citet{gema2024mmlu} re-annotate MMLU and
find 6.49\% of items contain errors. At the design level,
\citet{vendrow2025reliability} motivate ``platinum benchmarks''
curated to minimise label noise, \citet{reuel2024betterbench}
aggregate 46 best-practice criteria for benchmark construction,
and contamination-resistant suites such as LiveCodeBench
\citep{jain2024livecodebench} update problems continuously.
\citet{huang2026deepfactcoevolvingbenchmarksagents} similarly treat
benchmarks as revisable artifacts, using an audit-then-score process
to update DeepFact-Bench labels and rationales before scoring models. Most
directly related, \citet{zhu2025rigorous} formalise the validity
threats facing agentic benchmarks and propose an Agentic Benchmark
Checklist of best practices spanning task validity, outcome
validity, and benchmark reporting; their assessment, however, is
performed manually by domain experts on a fixed set of benchmarks.
These efforts proceed benchmark-by-benchmark, with critique
encoded as expert prose rather than structured records. We instead
define an executable audit protocol that emits a stable finding
schema under a single rubric and applies uniformly across
heterogeneous benchmark families, turning ad hoc critique into a
reproducible measurement.

\subsection{Benchmarking of AI Agents and LLMs}

Agentic and LLM benchmarks have proliferated across domains. In
coding and software engineering,
HumanEval \citep{chen2021humaneval} and
SWE-bench \citep{jimenez2024swebench} have been extended
by repository- and project-scale successors and
contamination-resistant variants
\citep{aleithan2024swebenchplus, deng2025swebenchpro,
miserendino2025swelancer, DBLP:conf/icml/Ouyang0AZHRM25, li2023can,
jain2024livecodebench}. Interactive agent benchmarks
exercise full execution environments and trajectories
\citep{liu2023agentbench, yao2025taubench, zhou2024webarena,
merrill2026terminalbench}, while multimodal agents extend
this to GUI and visual browsing tasks
\citep{xie2024osworld, he2024webvoyager}. Safety and
alignment benchmarks evaluate cyber capabilities, vulnerability
exploitation, and safety risks
\citep{zhang2024cybench, zhu2025cvebench, vidgen2024aisafety, kwon2025reasoniflargereasoningmodels}.
Other suites target advanced math, ML engineering, and
general-assistant tasks
\citep{mialon2023gaia, glazer2024frontiermath, nie2026dsgym}.
Aggregate evaluation frameworks such as HELM
\citep{liang2022helm} and BIG-bench
\citep{srivastava2023bigbench} standardize scenarios across many
benchmarks but treat the underlying benchmarks as trustworthy
inputs.

Agent--environment interaction, dynamic evaluation, and stateful
sandboxes expand the surface area for environment bugs, harness
flakiness, and narrow tests. Our framework targets this
multi-domain setting, producing audit-grade evidence per task under
a single rubric.

\subsection{From LLM-as-Judge to Agentic Audit}

Our auditor builds on the methodology of LLM-as-judge work but
moves from a single-call judgment to an agentic process.
\citet{zheng2023mtbench} demonstrate that strong LLM judges agree
with human raters above 80\% on open-ended chat evaluation, and
Chatbot Arena \citep{chiang2024arena} validates this
against crowdsourced preferences at scale. 
\citet{feuer2025judgenoise} document systematic failure modes such as
position, verbosity, and rubric-order biases that silently
undermine validity. These methods score model outputs given a fixed benchmark
via a single, stateless judgment. We instead score
benchmarks themselves using an agentic audit: the
auditor is given tools for filesystem inspection, code execution,
container introspection, and targeted retrieval, and proceeds
through multi-step investigation, the setting where harness
leaks, environment drift, and contradictory ground truth typically
first surface. Concurrent with our work,
\citet{tu2026benchguard} propose BenchGuard, which cross-verifies
benchmark artifacts via structured LLM protocols and reports issues
on two scientific benchmarks (ScienceAgentBench and a BixBench
subset). Our framework differs in two respects: the audit
pipeline is benchmark-agnostic and imposes no fixed preparation
format on the benchmark under test, and we evaluate at
substantially larger scale: 168 benchmarks spanning coding,
agentic, multimodal, and safety domains, versus two in their work, yielding a
cross-domain measurement of audit-grade evidence rather than a
per-benchmark study.

% !TEX root = ../neurips_2026.tex
\section{Conclusion}

In this work, we have demonstrated that current frontier benchmarks contain a wide range of issues of major severity. This calls for cyclically improving benchmarks as part of the standard development process. In the open-source software community, the maturation of systems gave rise to systematic testing and continuous improvement infrastructure precisely because complexity at scale makes issues inevitable, independent of the expertise of the contributors. AI benchmarks now face a very similar condition. We hope ABA can represent a step toward the analogous paradigm where an automated auditing framework can surface ambiguities, environment conflicts, and evaluation problems that domain experts may miss. We release all findings and tooling publicly.

\label{sec:conclusion}
\clearpage
% !TEX root = ../neurips_2026.tex
\begin{ack}
\end{ack}

% !TEX root = ../neurips_2026.tex
\small
\bibliographystyle{plainnat}
\bibliography{references}

\appendix

% !TEX root = ../neurips_2026.tex
\clearpage

\section{More Details on Task Audits}
\label{app:rubrics}

% \begin{figure}[h]
%   \centering
%   \begin{tcolorbox}[
%     enhanced, colback=white, colframe=black!55,
%     title={\textbf{Task-level audit rubric}\hfill
%            \normalfont\footnotesize severity 0--2 \,\textbar\,
%            subtype assigned post-hoc by classifier},
%     fonttitle=\bfseries, coltitle=white, colbacktitle=black!75,
%     boxrule=0.4pt, arc=2pt,
%     left=6pt, right=6pt, top=5pt, bottom=5pt,
%     fontupper=\footnotesize
%   ]
%   \begin{tabularx}{\linewidth}{@{}>{\raggedright\arraybackslash}p{2.6cm}@{\hspace{6pt}}>{\raggedright\arraybackslash}X@{}}
%     \textcolor{rubricA}{\textbf{Ambiguity}} &
%       \chip{rubricA}{A1}{API}\chip{rubricA}{A2}{Misleading}\chip{rubricA}{A3}{Format}\chip{rubricA}{A4}{Temporal}\chip{rubricA}{A5}{Destruction}\chip{rubricA}{A6}{TestMech}\chip{rubricA}{A7}{Cleanup} \\[3pt]
%     \textcolor{rubricE}{\textbf{Environment}} &
%       \chip{rubricE}{E1}{Tool}\chip{rubricE}{E2}{Resources}\chip{rubricE}{E3}{Filesystem}\chip{rubricE}{E4}{Harness}\chip{rubricE}{E5}{External} \\[3pt]
%     \textcolor{rubricT}{\textbf{Test Quality}} &
%       \chip{rubricT}{T1}{Narrow}\chip{rubricT}{T2}{Coverage}\chip{rubricT}{T3}{Target}\chip{rubricT}{T4}{Flaky}\chip{rubricT}{T5}{Gameable} \\
%   \end{tabularx}
%   \end{tcolorbox}
%   \caption{Task-level rubric overview: three categories,
%     17 subtypes, severity 0--2.}
%   \label{fig:rubric_task}
% \end{figure}

\subsection{Severity scale and calibration}
\label{app:rubrics:severity}

Severity is operationalised by a single thought experiment: the
auditor imagines handing the prompt --- and only the prompt ---
to five experienced developers in the task's domain, and asks
whether most could reliably produce a solution the tests accept.
The score on each axis answers that question:

\begin{center}
\begin{tabularx}{\linewidth}{@{}c l X@{}}
\toprule
\textbf{Level} & \textbf{Label} & \textbf{Meaning} \\
\midrule
0 & None  & The prompt is clear and sufficient. The tests fairly evaluate what the prompt asks for. Any failure is attributable to the agent's own mistakes or skill gaps. \\
1 & Minor & A discoverable gap. Reading only the prompt, a competent developer in the relevant domain can reach the expected solution --- but not immediately. The bridge requires deliberate work: consulting git history, a known API, reference docs, or reasoning through which of a few standard approaches the tests likely want. Some developers will cross it, others will pick a different reasonable interpretation that the tests reject. \\
2 & Major & It is almost impossible to understand what you are being asked to do without further information. The prompt omits or obscures key details that no amount of domain expertise can compensate for. The tests are too narrow, too broad, or look for something different from what the task description asks. Information may be missing, contradictory, or actively misleading. \\
\bottomrule
\end{tabularx}
\end{center}

\paragraph{Ambiguity vs.\ intrinsic difficulty.} The calibration
deliberately separates the two. Two corner cases both score 0,
by design: (i) a well-designed but hard task --- developers
understand what is being asked but find the work itself
difficult --- because the failure mode is skill, time, or
insight, not interpretation; and (ii) a vague prompt whose
expected approach is domain-standard, because domain knowledge
bridges the gap without deliberate effort. A non-zero severity
therefore requires that the failure mode be \emph{interpretive}:
a competent reader, given the prompt alone, could not reliably
determine what passing means.

\section{Audited benchmark portfolio}
\label{app:portfolio}

This appendix details the two selection sources that produce
the audited portfolio summarized in \S\ref{sec:audit:experiment_setup}:
(i)~the \emph{frontier release-report consensus}
(Appendix~\ref{app:portfolio:frontier}), which surfaces benchmarks
that frontier model developers themselves treat as load-bearing
evidence of capability, and (ii)~the \emph{NeurIPS-track sampling}
pass (Appendix~\ref{app:portfolio:neurips2025}), which classifies
the most recent NeurIPS Datasets \& Benchmarks Track papers and
retains those that fall within the audit's nine in-scope domains.
Appendix~\ref{app:portfolio:inventory} gives the full
per-benchmark inventory.

\subsection{Frontier Benchmarks}
\label{app:portfolio:frontier}

\begin{figure}[t!]
  \centering
  \includegraphics[width=\linewidth]{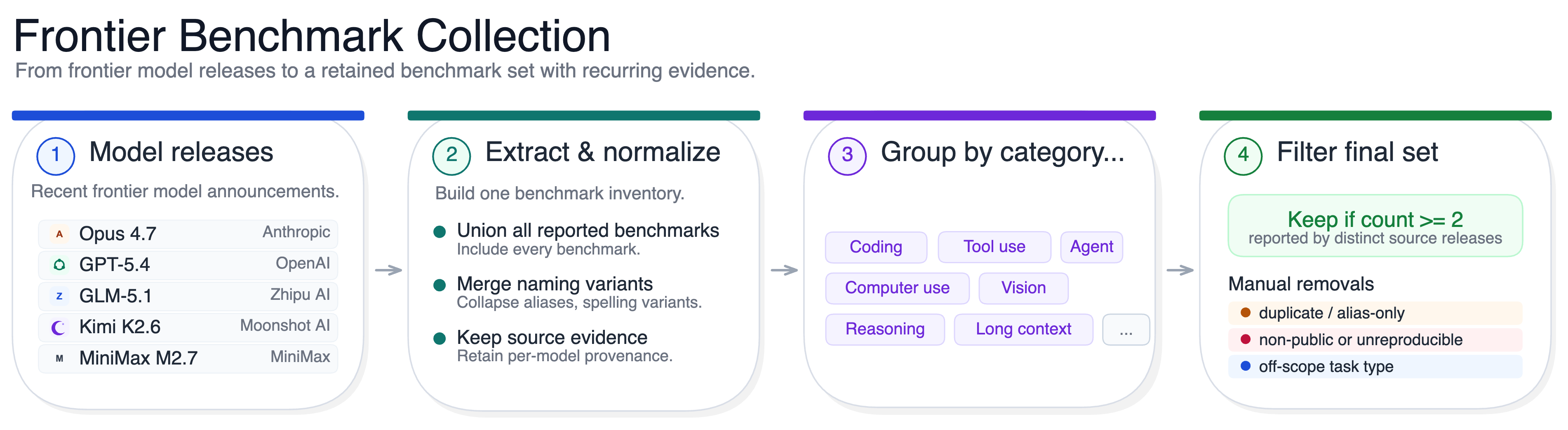}
  \caption{Frontier benchmark collection funnel. Five recent
  frontier model release reports (Opus~4.7, GPT-5.4, GLM-5.1,
  Kimi~K2.6, MiniMax~M2.7) are unioned, normalized, and filtered
  to the set of benchmarks cited by at least two releases, with
  manual removals for duplicates and off-scope task types.}
  \label{fig:portfolio:funnel}
\end{figure}

We assemble the frontier portion of the portfolio from the
benchmark tables published in five recent frontier model release
reports: Anthropic Opus~4.7~\citep{anthropic2026opus47}, OpenAI
GPT-5.4~\citep{openai2026gpt54}, Zhipu
GLM-5.1~\citep{glm5team2026glm5}, Moonshot
Kimi~K2.6~\citep{moonshot2026kimik26}, and MiniMax
M2.7~\citep{minimax2026m27}. The pipeline runs in four
passes (Fig.~\ref{fig:portfolio:funnel}).
\emph{(1)~Extraction.} For each release report, we pull every
benchmark named in the headline capability tables, retaining
the cited subset (e.g.\ Verified, Hard) and the model-vs-baseline
configuration where reported.
\emph{(2)~Alias normalization.} We collapse aliases and suite
variants -- e.g.\ ``SWE-bench Verified'' and ``SWEBench-V''
to a single canonical entry, and the year-tagged AIME variants
to a single AIME family -- while preserving per-model provenance
so we can audit which release surfaced each benchmark.
\emph{(3)~$\geq 2$-report intersection.} We retain only
benchmarks cited by at least two distinct release reports. This
operationalises the assumption that a benchmark which multiple
frontier developers independently chose for their report card
is load-bearing evidence of capability, not a one-off pick.
\emph{(4)~Manual scoping.} We then drop entries that fall
outside the audit's domain scope
and consolidate residual duplicates surfaced during inspection.
The surviving benchmarks form the frontier source set audited
in this paper.

\subsection{NeurIPS 2025 Datasets \& Benchmarks Track}
\label{app:portfolio:neurips2025}

We pair the frontier source with a sweep of the most recent
NeurIPS Datasets \& Benchmarks (D\&B) Track. We start from every
accepted NeurIPS 2025 D\&B paper that introduces a benchmark
(pure dataset releases without an evaluation protocol are
excluded). We classify each paper into one of sixteen candidate
domains using the paper's stated task, modality, and grader
(LLM-assisted classification with manual spot checks on
boundary cases). We then retain only papers whose domain falls
within the nine in-scope domains used throughout the paper --
science / expert reasoning, multimodal / vision reasoning,
professional / economic work, interactive / agentic / tool use,
coding / SWE / terminal, medical / clinical, math / formal
reasoning, retrieval / RAG / search, and safety / alignment.
No popularity- or citation-based ranking is applied: every
accepted benchmark whose domain matches the scope is audited.

\paragraph{Excluded candidate domains.}
The remaining seven candidate domains are dropped by the scoping
criterion of \S\ref{sec:audit:experiment_setup:benchmarks}. We list them
explicitly so the boundary of the audit is transparent.
\begin{itemize}
  \item \textbf{NLP / text.} Too broad and overlapping: ``NLP''
  as labelled by D\&B papers spans instruction following,
  summarisation, factual recall, classification, and translation,
  all of which fold into our science, reasoning, and retrieval
  buckets. Tracking it as a separate bucket would double-count
  tasks already audited elsewhere.
  \item \textbf{Creative generation.} Grading is inherently
  subjective and human-judgment-dominated; the audit rubric,
  which compares an instruction against a verifiable grader
  (test suite, gold answer, or deterministic check), has no
  meaningful operating point on benchmarks whose ground truth
  is ``a panel of humans liked this output''.
  \item \textbf{Evaluation methodology.} Auditing an
  evaluation-methodology benchmark with our auditor would be
  circular -- the artifact under test is itself an evaluation
  protocol, so the rubric would be measuring a property of the
  system that is supposed to instantiate the rubric.
  \item \textbf{Audio / speech.} The system under test is
  typically a specialized speech stack (ASR, TTS, audio
  classifiers) rather than a general-purpose LLM or LLM-driven
  agent; generalist audit findings do not transfer to that
  pipeline, and the rubric's grading-discipline items do not
  match the failure modes of speech evaluation.
  \item \textbf{Video understanding.} Same reason as
  audio/speech -- the system under test is a specialised video
  model and the relevant failure modes (temporal grounding,
  frame sampling, codec assumptions) lie outside the rubric's
  scope. Image-and-text multimodal evaluation is retained
  under multimodal / vision reasoning because those benchmarks
  target general-purpose multimodal LLMs.
  \item \textbf{Embodied 3D.} The system under test is an
  embodied agent operating against simulators or physical
  robots; the grading and environment-specification surface area
  is dominated by simulator parametrisation and physics, neither
  of which the audit rubric speaks to.
  \item \textbf{Remote sensing.} Specialised satellite/imagery
  stack with bespoke pre-processing pipelines and label
  conventions; like audio/speech, generalist audit findings do
  not transfer, and the system under test is rarely a frontier
  LLM.
\end{itemize}

\subsection{Per-Benchmark Overview}
\label{app:portfolio:inventory}

% !TEX root = ../neurips_2026.tex
% Per-benchmark overview: total tasks, tasks audited, and the count of
% audited tasks flagged with at least one major / minor finding by the
% static auditor. Generated from
% \texttt{draft\_content/all\_benchmarks.md} via
% \texttt{draft\_figures/scripts/per\_benchmark\_overview.py}.
% Requires \texttt{\textbackslash usepackage\{longtable\}} in the
% preamble of \texttt{neurips\_2026.tex}.

\small
\setlength{\LTleft}{\fill}
\setlength{\LTright}{\fill}
\setlength{\tabcolsep}{4pt}
\begin{longtable}{llrrrr}
  \caption{Per-benchmark overview across the full 168 benchmarks (static audit). \textbf{Tasks} is the total
  number of tasks discovered in the benchmark; \textbf{Audited} is the
  number of those tasks audited by the static pipeline; \textbf{Major}
  and \textbf{Minor} are the counts of audited tasks that received at
  least one finding at the corresponding severity. Rows are grouped by
  domain.}
  \label{tab:appendix:per_benchmark_overview} \\
  \toprule
  \textbf{Benchmark} & \textbf{Source} & \textbf{Tasks} & \textbf{Audited} & \textbf{Major} & \textbf{Minor} \\
  \midrule
  \endfirsthead
  \multicolumn{6}{l}{\emph{Table~\ref{tab:appendix:per_benchmark_overview} continued from previous page.}} \\
  \toprule
  \textbf{Benchmark} & \textbf{Source} & \textbf{Tasks} & \textbf{Audited} & \textbf{Major} & \textbf{Minor} \\
  \midrule
  \endhead
  \midrule
  \multicolumn{6}{r}{\emph{continued on next page}} \\
  \endfoot
  \bottomrule
  \endlastfoot
    \multicolumn{6}{l}{\textit{Science / expert reasoning}} \\
    \midrule
    \texttt{astrovisbench} & {\scriptsize\href{https://github.com/SebaJoe/AstroVisBench}{\texttt{AstroVisBench}}} &     432 &     200 &    91 &    69 \\
    \texttt{atmossci\_bench} & {\scriptsize\href{https://github.com/Relaxed-System-Lab/AtmosSci-Bench}{\texttt{AtmosSci-Bench}}} & 2{,}641 &     200 &    39 &    37 \\
    \texttt{beyond\_chemical\_qa} & {\scriptsize\href{https://github.com/IDEA-XL/ChemCoTBench}{\texttt{ChemCoTBench}}} & 1{,}485 &     500 &   151 &    67 \\
    \texttt{cellverse} & {\scriptsize\href{https://github.com/zfkarl/CellVerse}{\texttt{CellVerse}}} & 1{,}822 &     200 &    74 &    53 \\
    \texttt{chemx} & {\scriptsize\href{https://github.com/ai-chem/ChemX}{\texttt{ChemX}}} &      10 &      10 &     5 &     3 \\
    \texttt{fgbench} & {\scriptsize\href{https://github.com/xuanliugit/FGBench}{\texttt{FGBench}}} & 7{,}146 &     100 &     9 &     7 \\
    \texttt{gpqa-diamond} & {\scriptsize\href{https://github.com/idavidrein/gpqa}{\texttt{gpqa}}} &     198 &     198 &     5 &     9 \\
    \texttt{humanitys\_last\_exam\_hle} & {\scriptsize\href{https://github.com/centerforaisafety/hle}{\texttt{hle}}} & 2{,}500 &     500 &   137 &    69 \\
    \texttt{phybench} & {\scriptsize\href{https://github.com/phybench-official/phybench}{\texttt{phybench}}} &     500 &     200 &    38 &    36 \\
    \texttt{physgym} & {\scriptsize\href{https://github.com/principia-ai/PhysGym}{\texttt{PhysGym}}} &      97 &      97 &     7 &    11 \\
    \texttt{physics} & {\scriptsize\href{https://github.com/Zhengsh123/PHYSICS}{\texttt{PHYSICS}}} & 2{,}000 &     200 &    64 &    27 \\
    \texttt{qcircuitbench} & {\scriptsize\href{https://github.com/EstelYang/QCircuitBench}{\texttt{QCircuitBench}}} &      28 &      28 &     7 &     3 \\
    \texttt{scigym\_neurips} & {\scriptsize\href{https://github.com/h4duan/scigym-neurips}{\texttt{scigym-neurips}}} &     350 &     200 &     9 &    48 \\
    \texttt{sfe} & {\scriptsize\href{https://github.com/PrismaX-Team/sfe}{\texttt{sfe}}} & 1{,}660 &     100 &    35 &    29 \\
    \texttt{supergpqa} & {\scriptsize\href{https://github.com/SuperGPQA/SuperGPQA}{\texttt{SuperGPQA}}} & 26{,}529 &     500 &   142 &    75 \\
    \midrule
    \multicolumn{6}{l}{\textit{Multimodal / vision reasoning}} \\
    \midrule
    \texttt{agmmu} & {\scriptsize\href{https://github.com/AgMMU/AgMMU}{\texttt{AgMMU}}} &     772 &     200 &   128 &    15 \\
    \texttt{anomalycot} & {\scriptsize\href{https://github.com/Zhaolutuan/AnomalyCoT}{\texttt{AnomalyCoT}}} & 7{,}916 &     200 &    63 &    80 \\
    \texttt{blink\_twice} & {\scriptsize\href{https://github.com/PicoTrex/BLINK-Twice}{\texttt{BLINK-Twice}}} &     899 &     200 &    69 &    35 \\
    \texttt{bmmr} & {\scriptsize\href{https://github.com/WooooDyy/BMMR}{\texttt{BMMR}}} & 40{,}916 &     200 &    48 &    31 \\
    \texttt{capability} & {\scriptsize\href{https://github.com/ali-vilab/CAPability}{\texttt{CAPability}}} & 11{,}631 &     200 &     1 &    12 \\
    \texttt{chartmuseum} & {\scriptsize\href{https://github.com/Liyan06/ChartMuseum}{\texttt{ChartMuseum}}} & 1{,}162 &     200 &     3 &    14 \\
    \texttt{choice} & {\scriptsize\href{https://github.com/ShawnAn-WHU/CHOICE}{\texttt{CHOICE}}} & 10{,}507 &     200 &    27 &    17 \\
    \texttt{cmm} & {\scriptsize\href{https://github.com/DAMO-NLP-SG/CMM}{\texttt{CMM}}} & 2{,}400 &     200 &     1 &    12 \\
    \texttt{colorbench} & {\scriptsize\href{https://github.com/tianyi-lab/ColorBench}{\texttt{ColorBench}}} & 5{,}885 &     100 &     7 &     7 \\
    \texttt{coralvqa} & {\scriptsize\href{https://huggingface.co/datasets/CoralReefData/CoralVQA}{\texttt{CoralVQA}}} & 50{,}927 &     100 &    15 &    18 \\
    \texttt{disasterm3} & {\scriptsize\href{https://github.com/junjue-wang/disasterm3}{\texttt{disasterm3}}} & 30{,}042 &     500 &   203 &    56 \\
    \texttt{drive\_mllm} & {\scriptsize\href{https://github.com/XiandaGuo/Drive-MLLM}{\texttt{Drive-MLLM}}} & 4{,}666 &     100 &    17 &    25 \\
    \texttt{face\_human\_bench} & {\scriptsize\href{https://github.com/lxq1000/Face-Human-Bench}{\texttt{Face-Human-Ben…}}} & 3{,}600 &     200 &     0 &     5 \\
    \texttt{finegrain\_eval} & {\scriptsize\href{https://github.com/khayes95/FineGRAIN_Eval}{\texttt{FineGRAIN\_Eval}}} &     760 &     200 &   158 &    35 \\
    \texttt{fire360dataset} & {\scriptsize\href{https://uofi.app.box.com/v/fire360dataset}{\texttt{fire360dataset}}} &       5 &       5 &     5 &     0 \\
    \texttt{from\_flatland\_to\_space} & {\scriptsize\href{https://github.com/LogosRoboticsGroup/SPAR}{\texttt{SPAR}}} & 7{,}211 &     500 &    26 &   160 \\
    \texttt{hyperphantasia} & {\scriptsize\href{https://github.com/AIF4S/Hyperphantasia}{\texttt{Hyperphantasia}}} & 1{,}200 &     200 &     0 &     0 \\
    \texttt{infochartqa} & {\scriptsize\href{https://github.com/thu-vis/InfoChartQA}{\texttt{InfoChartQA}}} & 58{,}857 &     200 &    60 &    20 \\
    \texttt{ir3d\_bench} & {\scriptsize\href{https://github.com/LiuHengyu321/IR3D-Bench}{\texttt{IR3D-Bench}}} & 15{,}000 &     200 &   173 &    19 \\
    \texttt{mimeqa} & {\scriptsize\href{https://github.com/MIT-MI/MimeQA}{\texttt{MimeQA}}} &     806 &     200 &     1 &     4 \\
    \texttt{mllm\_isu} & {\scriptsize\href{https://github.com/1012537710/MLLM-ISU}{\texttt{MLLM-ISU}}} & 3{,}000 &     200 &   138 &    24 \\
    \texttt{mm\_opera} & {\scriptsize\href{https://github.com/MM-OPERA-Bench/MM-OPERA}{\texttt{MM-OPERA}}} & 11{,}497 &     200 &    37 &    52 \\
    \texttt{mmcsbench} & {\scriptsize\href{https://github.com/zhangjinCV/MMCSBench}{\texttt{MMCSBench}}} & 79{,}794 &     200 &   141 &    32 \\
    \texttt{mme} & {\scriptsize\href{https://huggingface.co/datasets/darkyarding/MME}{\texttt{MME}}} & 2{,}374 &     200 &     6 &     6 \\
    \texttt{mme\_videoocr} & {\scriptsize\href{https://github.com/FrankYang-17/MME-VideoOCR}{\texttt{MME-VideoOCR}}} & 2{,}000 &     200 &    21 &    10 \\
    \texttt{mmla} & {\scriptsize\href{https://github.com/thuiar/MMLA}{\texttt{MMLA}}} & 15{,}708 &     200 &    79 &    22 \\
    \texttt{mmlongbench} & {\scriptsize\href{https://github.com/edinburghnlp/mmlongbench}{\texttt{mmlongbench}}} & 7{,}801 &     500 &    25 &    64 \\
    \texttt{mmmu\_pro} & {\scriptsize\href{https://github.com/MMMU-Benchmark/MMMU/tree/main/mmmu-pro}{\texttt{mmmu-pro}}} & 1{,}730 &     501 &    62 &    38 \\
    \texttt{mmpb} & {\scriptsize\href{https://github.com/MMPB-Benchmark/MMPB}{\texttt{MMPB}}} & 10{,}017 &     200 &    16 &    35 \\
    \texttt{mmperspective} & {\scriptsize\href{https://github.com/yunlong10/MMPerspective}{\texttt{MMPerspective}}} & 5{,}083 &     200 &     4 &     7 \\
    \texttt{ocrbench\_v2} & {\scriptsize\href{https://github.com/Yuliang-Liu/MultimodalOCR}{\texttt{MultimodalOCR}}} & 10{,}000 &     500 &   138 &    69 \\
    \texttt{pacbench} & {\scriptsize\href{https://github.com/Atharva-Gundawar/pacbench}{\texttt{pacbench}}} & 11{,}881 &     200 &    74 &    91 \\
    \texttt{robo2vlm} & {\scriptsize\href{https://github.com/KeplerC/robo2VLM}{\texttt{robo2VLM}}} & 6{,}676 &     200 &   176 &    11 \\
    \texttt{seephys\_project} & {\scriptsize\href{https://github.com/SeePhys/seephys-project}{\texttt{seephys-project}}} & 2{,}000 &     200 &    32 &    23 \\
    \texttt{spuriverse} & {\scriptsize\href{https://github.com/Anderson-Lee-Git/SpuriVerse}{\texttt{SpuriVerse}}} &     124 &     124 &    21 &    39 \\
    \texttt{uve} & {\scriptsize\href{https://github.com/bytedance/UVE}{\texttt{UVE}}} & 4{,}221 &     200 &    11 &    12 \\
    \texttt{vlmevalkit\_rbench\_v} & {\scriptsize\href{https://github.com/CHEN-Xinsheng/VLMEvalKit_RBench-V}{\texttt{VLMEvalKit\_RBe…}}} &     803 &     200 &    45 &     7 \\
    \midrule
    \multicolumn{6}{l}{\textit{Professional / economic work}} \\
    \midrule
    \texttt{concordia} & {\scriptsize\href{https://github.com/google-deepmind/concordia}{\texttt{concordia}}} &      18 &      18 &    11 &     1 \\
    \texttt{dabstep} & {\scriptsize\href{https://huggingface.co/datasets/adyen/DABstep}{\texttt{DABstep}}} &     450 &     450 &   120 &   107 \\
    \texttt{deepfund} & {\scriptsize\href{https://github.com/HKUSTDial/DeepFund}{\texttt{DeepFund}}} &      13 &      13 &    11 &     0 \\
    \texttt{finance\_agent} & {\scriptsize\href{https://github.com/vals-ai/finance-agent}{\texttt{finance-agent}}} &      50 &      50 &     7 &     9 \\
    \texttt{gdpval\_aa} & {\scriptsize\href{https://huggingface.co/datasets/openai/gdpval}{\texttt{gdpval}}} &     220 &     220 &    80 &    95 \\
    \texttt{multimodal\_oil\_gas\_benchmark} & {\scriptsize\href{https://github.com/climate-nlp/multimodal-oil-gas-benchmark}{\texttt{multimodal-oil…}}} &     353 &     200 &   144 &    25 \\
    \texttt{officeqa\_pro} & {\scriptsize\href{https://github.com/databricks/officeqa}{\texttt{officeqa}}} &     133 &     133 &    42 &    38 \\
    \texttt{steer\_me} & {\scriptsize\href{https://huggingface.co/datasets/narunraman/steer_me}{\texttt{steer\_me}}} &      76 &      76 &    25 &     8 \\
    \midrule
    \multicolumn{6}{l}{\textit{Interactive / agentic / tool-use}} \\
    \midrule
    \texttt{agentic\_benchmarks} & {\scriptsize\href{https://github.com/uiuc-kang-lab/agentic-benchmarks}{\texttt{agentic-benchm…}}} &     269 &     200 &    15 &    40 \\
    \texttt{agentif} & {\scriptsize\href{https://github.com/THU-KEG/AgentIF}{\texttt{AgentIF}}} &     707 &     200 &    93 &    63 \\
    \texttt{agentrecbench} & {\scriptsize\href{https://github.com/Nghia9211/AgentRecBench}{\texttt{AgentRecBench}}} & 1{,}499 &     200 &     1 &     0 \\
    \texttt{ai\_agent\_privacy} & {\scriptsize\href{https://github.com/facebookresearch/ai-agent-privacy}{\texttt{ai-agent-priva…}}} &     246 &     200 &    42 &    61 \\
    \texttt{ale\_bench} & {\scriptsize\href{https://github.com/SakanaAI/ALE-Bench}{\texttt{ALE-Bench}}} &      40 &      40 &     0 &     0 \\
    \texttt{factorio\_learning\_environment} & {\scriptsize\href{https://github.com/JackHopkins/factorio-learning-environment}{\texttt{factorio-learn…}}} &      30 &      30 &     2 &     0 \\
    \texttt{livevqa} & {\scriptsize\href{https://github.com/fumingyang2004/LIVEVQA}{\texttt{LIVEVQA}}} & 2{,}384 &     200 &    85 &    33 \\
    \texttt{llm\_speedrunner} & {\scriptsize\href{https://github.com/facebookresearch/llm-speedrunner}{\texttt{llm-speedrunner}}} &      20 &      20 &     2 &     5 \\
    \texttt{medagentboard} & {\scriptsize\href{https://github.com/yhzhu99/MedAgentBoard}{\texttt{MedAgentBoard}}} & 2{,}279 &     200 &    20 &    38 \\
    \texttt{mineanybuild} & {\scriptsize\href{https://github.com/MineAnyBuild/MineAnyBuild}{\texttt{MineAnyBuild}}} & 4{,}616 &     200 &    71 &    55 \\
    \texttt{mle\_dojo} & {\scriptsize\href{https://github.com/MLE-Dojo/MLE-Dojo}{\texttt{MLE-Dojo}}} &     202 &     200 &    47 &    42 \\
    \texttt{mlrbench} & {\scriptsize\href{https://github.com/chchenhui/mlrbench}{\texttt{mlrbench}}} &     201 &     200 &     9 &    40 \\
    \texttt{mlrc\_bench} & {\scriptsize\href{https://github.com/yunx-z/MLRC-Bench}{\texttt{MLRC-Bench}}} &       8 &       8 &     3 &     4 \\
    \texttt{opencaptchaworld} & {\scriptsize\href{https://github.com/MetaAgentX/OpenCaptchaWorld}{\texttt{OpenCaptchaWor…}}} &     463 &     200 &    54 &    25 \\
    \texttt{osworld\_g} & {\scriptsize\href{https://github.com/xlang-ai/osworld-g}{\texttt{osworld-g}}} &     564 &     200 &    20 &    33 \\
    \texttt{osworld\_verified} & {\scriptsize\href{https://github.com/xlang-ai/OSWorld}{\texttt{OSWorld}}} &     369 &     369 &    60 &    75 \\
    \texttt{real} & {\scriptsize\href{https://github.com/agi-inc/REAL}{\texttt{REAL}}} &     112 &     112 &    17 &    22 \\
    \texttt{recbench} & {\scriptsize\href{https://github.com/Jyonn/RecBench}{\texttt{RecBench}}} &      15 &      15 &     4 &     7 \\
    \texttt{t1} & {\scriptsize\href{https://github.com/CapitalOne-Research/T1}{\texttt{T1}}} & 9{,}000 &     200 &    54 &    53 \\
    \texttt{tau2\_bench\_telecom} & {\scriptsize\href{https://github.com/sierra-research/tau2-bench}{\texttt{tau2-bench}}} &     114 &     114 &     0 &     5 \\
    \texttt{the\_agent\_company} & {\scriptsize\href{https://github.com/TheAgentCompany/TheAgentCompany}{\texttt{TheAgentCompany}}} &     175 &     100 &    34 &    27 \\
    \texttt{toolathlon} & {\scriptsize\href{https://github.com/hkust-nlp/Toolathlon}{\texttt{Toolathlon}}} &     108 &     108 &    44 &    47 \\
    \texttt{wasp} & {\scriptsize\href{https://github.com/facebookresearch/wasp}{\texttt{wasp}}} &      21 &      21 &     4 &     4 \\
    \midrule
    \multicolumn{6}{l}{\textit{Coding / SWE / terminal}} \\
    \midrule
    \texttt{aider-polyglot} & {\scriptsize\href{https://github.com/Aider-AI/polyglot-benchmark}{\texttt{polyglot-bench…}}} &     225 &     225 &    77 &    56 \\
    \texttt{algotune} & {\scriptsize\href{https://github.com/oripress/AlgoTune}{\texttt{AlgoTune}}} &     154 &     154 &    21 &    36 \\
    \texttt{clever} & {\scriptsize\href{https://github.com/trishullab/clever}{\texttt{clever}}} &     161 &     161 &    66 &    16 \\
    \texttt{codeassistbench} & {\scriptsize\href{https://github.com/amazon-science/CodeAssistBench}{\texttt{CodeAssistBench}}} &     274 &     200 &    53 &    46 \\
    \texttt{core} & {\scriptsize\href{https://github.com/CoReBench/CoRe}{\texttt{CoRe}}} & 23{,}955 &     200 &     4 &    30 \\
    \texttt{cpret} & {\scriptsize\href{https://github.com/coldchair/CPRet}{\texttt{CPRet}}} & 19{,}806 &     200 &     1 &     4 \\
    \texttt{effibench\_x} & {\scriptsize\href{https://github.com/EffiBench/EffiBench-X}{\texttt{EffiBench-X}}} &     623 &     200 &     3 &    16 \\
    \texttt{frontier\_swe} & {\scriptsize\href{https://github.com/Proximal-Labs/frontier-swe}{\texttt{frontier-swe}}} &      17 &      17 &     4 &     4 \\
    \texttt{gso} & {\scriptsize\href{https://github.com/gso-bench/gso}{\texttt{gso}}} &     102 &     102 &     5 &    19 \\
    \texttt{icpc\_eval} & {\scriptsize\href{https://github.com/RUCAIBox/ICPC-Eval}{\texttt{ICPC-Eval}}} &     118 &     118 &    32 &     8 \\
    \texttt{ir\_optset} & {\scriptsize\href{https://github.com/yilingqinghan/IR-OptSet}{\texttt{IR-OptSet}}} & 170{,}564 &     200 &    30 &    31 \\
    \texttt{livecodebench\_pro} & {\scriptsize\href{https://github.com/GavinZhengOI/LiveCodeBench-Pro}{\texttt{LiveCodeBench-…}}} &     864 &     191 &     3 &     4 \\
    \texttt{llm4decompile} & {\scriptsize\href{https://github.com/albertan017/LLM4Decompile}{\texttt{LLM4Decompile}}} & 9{,}104 &     200 &    25 &    13 \\
    \texttt{multi\_swe\_bench} & {\scriptsize\href{https://github.com/multi-swe-bench/multi-swe-bench}{\texttt{multi-swe-bench}}} & 1{,}632 &     200 &    42 &    34 \\
    \texttt{parrot} & {\scriptsize\href{https://github.com/weAIDB/PARROT}{\texttt{PARROT}}} & 28{,}003 &     200 &    64 &    23 \\
    \texttt{researchcodebench} & {\scriptsize\href{https://github.com/PatrickHua/ResearchCodeBench}{\texttt{ResearchCodeBe…}}} &     212 &     200 &    14 &    19 \\
    \texttt{swe\_bench} & {\scriptsize\href{https://github.com/SWE-bench/SWE-bench}{\texttt{SWE-bench}}} &     500 &     500 &    21 &    36 \\
    \texttt{swe\_bench\_fork} & {\scriptsize\href{https://github.com/SWE-rebench/SWE-bench-fork}{\texttt{SWE-bench-fork}}} &     750 &     200 &    49 &    38 \\
    \texttt{swe\_bench\_live} & {\scriptsize\href{https://github.com/microsoft/SWE-bench-Live}{\texttt{SWE-bench-Live}}} & 1{,}887 &     200 &    70 &    37 \\
    \texttt{swe\_bench\_multilingual} & {\scriptsize\href{https://huggingface.co/datasets/SWE-bench/SWE-bench_Multilingual}{\texttt{SWE-bench\_Mult…}}} &     300 &     300 &    12 &    30 \\
    \texttt{swe-bench-pro} & {\scriptsize\href{https://github.com/scaleapi/SWE-bench_Pro-os}{\texttt{SWE-bench\_Pro-…}}} &     731 &     731 &    85 &   142 \\
    \texttt{tb2} & {\scriptsize\href{https://github.com/laude-institute/terminal-bench-2}{\texttt{terminal-bench…}}} &      89 &      89 &     5 &    19 \\
    \texttt{tf\_bench} & {\scriptsize\href{https://github.com/SecurityLab-UCD/TF-Bench}{\texttt{TF-Bench}}} &     188 &     188 &     5 &    23 \\
    \texttt{webgen\_bench} & {\scriptsize\href{https://github.com/mnluzimu/WebGen-Bench}{\texttt{WebGen-Bench}}} &     101 &      65 &    16 &    32 \\
    \midrule
    \multicolumn{6}{l}{\textit{Medical / clinical}} \\
    \midrule
    \texttt{cgbench} & {\scriptsize\href{https://github.com/owencqueen/cgbench}{\texttt{cgbench}}} & 2{,}514 &     200 &   179 &    11 \\
    \texttt{clinbench} & {\scriptsize\href{https://github.com/ismaelvillanuevamiranda/ClinBench}{\texttt{ClinBench}}} & 2{,}876 &     200 &   195 &     2 \\
    \texttt{clinicallab} & {\scriptsize\href{https://github.com/WeixiangYAN/ClinicalLab}{\texttt{ClinicalLab}}} &      16 &      16 &    13 &     3 \\
    \texttt{cxreasonbench} & {\scriptsize\href{https://github.com/ttumyche/CXReasonBench}{\texttt{CXReasonBench}}} &      12 &      12 &     2 &     6 \\
    \texttt{drvd\_bench} & {\scriptsize\href{https://github.com/Jerry-Boss/DrVD-Bench}{\texttt{DrVD-Bench}}} & 7{,}789 &     200 &    34 &    62 \\
    \texttt{endobench} & {\scriptsize\href{https://github.com/CUHK-AIM-Group/EndoBench}{\texttt{EndoBench}}} & 6{,}832 &     500 &    17 &    35 \\
    \texttt{ltd\_bench} & {\scriptsize\href{https://github.com/walktaster/LTD-Bench}{\texttt{LTD-Bench}}} &     183 &     183 &    19 &    27 \\
    \texttt{medchain} & {\scriptsize\href{https://github.com/ljwztc/MedChain}{\texttt{MedChain}}} & 2{,}362 &     500 &   411 &    53 \\
    \texttt{medmax} & {\scriptsize\href{https://github.com/Hritikbansal/medmax}{\texttt{medmax}}} & 9{,}467 &     200 &    64 &    28 \\
    \texttt{medsg\_bench} & {\scriptsize\href{https://github.com/Yuejingkun/MedSG-Bench}{\texttt{MedSG-Bench}}} & 9{,}630 &     200 &    65 &    44 \\
    \texttt{mtbbench} & {\scriptsize\href{https://github.com/bunnelab/MTBBench}{\texttt{MTBBench}}} &     573 &     500 &   108 &    76 \\
    \texttt{oralgpt} & {\scriptsize\href{https://github.com/isbrycee/OralGPT}{\texttt{OralGPT}}} & 1{,}069 &     500 &   100 &   118 \\
    \texttt{patientsim} & {\scriptsize\href{https://github.com/dek924/PatientSim}{\texttt{PatientSim}}} &       3 &       3 &     0 &     0 \\
    \texttt{smmile} & {\scriptsize\href{https://github.com/eth-medical-ai-lab/smmile}{\texttt{smmile}}} & 1{,}149 &     200 &    12 &    21 \\
    \texttt{tcm\_ladder} & {\scriptsize\href{https://github.com/orangeshushu/TCM-Ladder}{\texttt{TCM-Ladder}}} & 29{,}262 &     200 &   115 &    18 \\
    \texttt{thousand\_voices\_trauma} & {\scriptsize\href{https://huggingface.co/datasets/yenopoya/thousand-voices-trauma}{\texttt{thousand-voice…}}} &     500 &     200 &   119 &     0 \\
    \texttt{vivabench} & {\scriptsize\href{https://github.com/chy-chiu/vivabench}{\texttt{vivabench}}} &     990 &     200 &   113 &    30 \\
    \midrule
    \multicolumn{6}{l}{\textit{Math / formal reasoning}} \\
    \midrule
    \texttt{aime\_fix} & {\scriptsize\href{https://huggingface.co/datasets/HuggingFaceH4/aime_2024}{\texttt{aime\_2024}}} &      60 &      60 &     0 &     3 \\
    \texttt{connectomebench} & {\scriptsize\href{https://github.com/jffbrwn2/ConnectomeBench}{\texttt{ConnectomeBench}}} & 1{,}827 &     200 &     5 &    17 \\
    \texttt{hardmath2\_eval} & {\scriptsize\href{https://github.com/JamesRoggeveen/hardmath2_eval}{\texttt{hardmath2\_eval}}} &     215 &     200 &    68 &    34 \\
    \texttt{imoanswerbench} & {\scriptsize\href{https://github.com/google-deepmind/superhuman/tree/main/imobench}{\texttt{imobench}}} &     400 &     400 &    14 &    35 \\
    \texttt{ineqmath} & {\scriptsize\href{https://github.com/lupantech/ineqmath}{\texttt{ineqmath}}} &     300 &     300 &    21 &    24 \\
    \texttt{leanineqcomp} & {\scriptsize\href{https://github.com/haoyuzhao123/LeanIneqComp}{\texttt{LeanIneqComp}}} &     375 &     200 &     6 &     0 \\
    \texttt{lexicon\_neurips} & {\scriptsize\href{https://github.com/periklismant/lexicon_neurips}{\texttt{lexicon\_neurips}}} &     666 &     200 &    77 &    49 \\
    \texttt{matharena} & {\scriptsize\href{https://github.com/eth-sri/matharena}{\texttt{matharena}}} &     951 &     500 &     6 &     9 \\
    \texttt{natural\_reasoning} & {\scriptsize\href{https://huggingface.co/datasets/facebook/natural_reasoning}{\texttt{natural\_reason…}}} & 1{,}145{,}824 &     200 &   118 &    27 \\
    \texttt{oeis\_sequence\_benchmark} & {\scriptsize\href{https://github.com/ceodspspectrum/oeis-sequence-benchmark}{\texttt{oeis-sequence-…}}} & 1{,}000 &     200 &    30 &    19 \\
    \texttt{omega} & {\scriptsize\href{https://github.com/sunblaze-ucb/omega}{\texttt{omega}}} & 9{,}715 &     100 &    17 &     7 \\
    \texttt{realmath} & {\scriptsize\href{https://github.com/ethz-spylab/RealMath}{\texttt{RealMath}}} & 1{,}286 &     100 &     6 &     8 \\
    \texttt{reasoning\_gym} & {\scriptsize\href{https://github.com/open-thought/reasoning-gym}{\texttt{reasoning-gym}}} & 5{,}000 &     200 &     6 &    11 \\
    \texttt{solidgeo} & {\scriptsize\href{https://github.com/HarryYancy/SolidGeo}{\texttt{SolidGeo}}} & 3{,}113 &     200 &    13 &     8 \\
    \midrule
    \multicolumn{6}{l}{\textit{Retrieval / RAG / search}} \\
    \midrule
    \texttt{c\_seo\_bench} & {\scriptsize\href{https://github.com/parameterlab/c-seo-bench}{\texttt{c-seo-bench}}} & 1{,}921 &     500 &     1 &    11 \\
    \texttt{freshstack} & {\scriptsize\href{https://github.com/fresh-stack/freshstack}{\texttt{freshstack}}} &     672 &     500 &    72 &    77 \\
    \texttt{hawkbench} & {\scriptsize\href{https://github.com/qhjqhj00/HawkBench}{\texttt{HawkBench}}} & 1{,}600 &     200 &    56 &    65 \\
    \texttt{kgqagen} & {\scriptsize\href{https://github.com/liangliang6v6/KGQAGen}{\texttt{KGQAGen}}} & 10{,}787 &     100 &    11 &    20 \\
    \texttt{mind2web} & {\scriptsize\href{https://github.com/OSU-NLP-Group/Mind2Web-2}{\texttt{Mind2Web-2}}} &      10 &      10 &     2 &     3 \\
    \texttt{mmdocrag} & {\scriptsize\href{https://github.com/MMDocRAG/MMDocRAG}{\texttt{MMDocRAG}}} & 4{,}055 &     500 &   139 &   166 \\
    \texttt{ms\_bench} & {\scriptsize\href{https://github.com/ianeong/MS-Bench}{\texttt{MS-Bench}}} & 8{,}729 &     200 &    42 &    88 \\
    \texttt{raguard} & {\scriptsize\href{https://huggingface.co/datasets/UCSC-IRKM/RAGuard}{\texttt{RAGuard}}} & 2{,}648 &     200 &    54 &    36 \\
    \midrule
    \multicolumn{6}{l}{\textit{Safety / alignment}} \\
    \midrule
    \texttt{agent\_red\_teaming} & {\scriptsize\href{https://github.com/GraySwanAI/agent-red-teaming}{\texttt{agent-red-team…}}} &      44 &      44 &    12 &    16 \\
    \texttt{artificial\_hivemind} & {\scriptsize\href{https://github.com/liweijiang/artificial-hiveminds}{\texttt{artificial-hiv…}}} & 1{,}350 &     500 &   255 &    28 \\
    \texttt{backdoorllm} & {\scriptsize\href{https://github.com/bboylyg/BackdoorLLM}{\texttt{BackdoorLLM}}} & 4{,}200 &     500 &   222 &   141 \\
    \texttt{benchmarkcards} & {\scriptsize\href{https://github.com/SokolAnn/BenchmarkCards}{\texttt{BenchmarkCards}}} & 4{,}721 &     200 &   197 &     0 \\
    \texttt{benchrisk} & {\scriptsize\href{https://github.com/BenchRisk/BenchRisk}{\texttt{BenchRisk}}} &      27 &      27 &    26 &     1 \\
    \texttt{chasm} & {\scriptsize\href{https://github.com/Jingyi62/CHASM}{\texttt{CHASM}}} & 1{,}000 &     200 &    33 &    31 \\
    \texttt{datasir} & {\scriptsize\href{https://github.com/Fan-Mo-ZJU/DataSIR}{\texttt{DataSIR}}} & 1{,}647{,}501 &     200 &    92 &    71 \\
    \texttt{deceptionbench} & {\scriptsize\href{https://github.com/Aries-iai/DeceptionBench}{\texttt{DeceptionBench}}} &     150 &     150 &    12 &    83 \\
    \texttt{intermt} & {\scriptsize\href{https://github.com/cby-pku/InterMT}{\texttt{InterMT}}} & 1{,}948 &     200 &   139 &    47 \\
    \texttt{open\_unlearning} & {\scriptsize\href{https://github.com/locuslab/open-unlearning}{\texttt{open-unlearning}}} &      48 &      48 &     1 &     4 \\
    \texttt{os\_harm} & {\scriptsize\href{https://github.com/tml-epfl/os-harm}{\texttt{os-harm}}} &     110 &     110 &     8 &    12 \\
    \texttt{panel\_understanding\_and\_operation} & {\scriptsize\href{https://huggingface.co/datasets/Tele-AI-MAIL/Panel-Understanding-and-Operation}{\texttt{Panel-Understa…}}} & 7{,}711 &     200 &    43 &    42 \\
    \texttt{pasbench} & {\scriptsize\href{https://huggingface.co/datasets/Youliang/PaSBench}{\texttt{PaSBench}}} &     416 &     200 &    21 &    25 \\
    \texttt{polyguard} & {\scriptsize\href{https://github.com/AI-secure/PolyGuard}{\texttt{PolyGuard}}} & 128{,}376 &     200 &    29 &    22 \\
    \texttt{prefcleanbench} & {\scriptsize\href{https://github.com/deeplearning-wisc/PrefCleanBench}{\texttt{PrefCleanBench}}} &     116 &     116 &   115 &     0 \\
    \texttt{safevideo} & {\scriptsize\href{https://github.com/Archippe-m-arkip/safevideo}{\texttt{safevideo}}} &       1 &       1 &     0 &     0 \\
    \texttt{sage\_eval} & {\scriptsize\href{https://github.com/YuehHanChen/SAGE-Eval}{\texttt{SAGE-Eval}}} & 1{,}105 &     200 &     6 &    15 \\
    \texttt{secodeplt} & {\scriptsize\href{https://github.com/ucsb-mlsec/SeCodePLT}{\texttt{SeCodePLT}}} & 10{,}596 &     200 &   176 &     5 \\
    \texttt{submission\_cares} & {\scriptsize\href{https://github.com/XiaominLi1998/Submission-CARES}{\texttt{Submission-CAR…}}} & 18{,}477 &     200 &    53 &     6 \\
    \texttt{umu\_bench} & {\scriptsize\href{https://github.com/QDRhhhh/UMU-bench}{\texttt{UMU-bench}}} &     653 &     200 &   142 &    37 \\
    \texttt{video\_safetybench} & {\scriptsize\href{https://github.com/flageval-baai/Video-SafetyBench}{\texttt{Video-SafetyBe…}}} &       4 &       4 &     2 &     0 \\
    \texttt{vmdt} & {\scriptsize\href{https://github.com/sunblaze-ucb/VMDT}{\texttt{VMDT}}} & 13{,}880 &     200 &    37 &    26 \\
\end{longtable}

\section{Auditor implementation}
\label{app:implementation}

This appendix documents the audit pipeline's output schema and
the reproducibility artefacts released alongside the paper.

\subsection{Output schema}
\label{app:schema}

Both audits emit findings under one record:
\texttt{\{finding\_id, category, subtype, severity, claim,
why\_it\_matters, evidence: [\{path, note\}], suggested\_fix\}}.
Benchmark-level records carry the
auditor-assigned subtype.

\subsection{Evidence collector prompts}
\label{app:collector_prompts}

Before any auditing happens, an evidence-collection agent is given
the prompt template in Listing~\ref{lst:collect:prompt} and asked
to produce the structured per-task artifact bundles that the
auditor later consumes (see
\S\ref{sec:framework:auditing_protocol}). The template runs in two
phases: Phase~1 ensures the benchmark materials are available
locally (cloning the repository, fetching the paper, and populating
the data directory), and Phase~2 writes and executes a Python
collector script that emits an \texttt{ArtifactManifest} and
per-task \texttt{TaskConfig} files. Tokens in square brackets
(e.g.\ \texttt{[BENCHMARK\_NAME]},
\texttt{[BENCHMARK\_DATA\_DIR]}, \texttt{[COLLECTOR\_PATH]}) are
substituted at render time from the
\texttt{EvidenceCollectionConfig}; the rest of the prompt is fixed.

\begin{tcblisting}{
  enhanced, breakable,
  colback=gray!3, colframe=black!55,
  title={Listing 1: Evidence collector prompt template},
  fonttitle=\bfseries, coltitle=white, colbacktitle=black!75,
  boxrule=0.4pt, arc=2pt,
  left=6pt, right=6pt, top=4pt, bottom=4pt,
  listing only,
  label={lst:collect:prompt},
  listing options={
    basicstyle=\ttfamily\scriptsize,
    breaklines=true,
    breakatwhitespace=true,
    columns=fullflexible,
    keepspaces=true,
    showstringspaces=false,
    upquote=true,
  },
}
You are an evidence collector for bench-audit.

## Phase 1: Ensure benchmark materials are available

Before collecting evidence, make sure the benchmark data is available locally.
The benchmark directory is organized as:
```
<benchmark_dir>/
|-- repo/    <- cloned repository (benchmark_repo_dir)
|-- data/    <- downloaded/processed dataset (benchmark_data_dir)
`-- paper.pdf <- benchmark paper (if available)
```

### Step 1a: Clone the repository (if needed)
If `benchmark_repo_dir` is provided and is empty or does not exist, and `code_url`
is available, clone the repository into `benchmark_repo_dir`.

### Step 1b: Download the paper (if needed)
If there is no paper PDF in the benchmark directory, try to find and download it.
Check the repo's README for an arXiv link or paper URL. If found, download the
PDF to the parent of benchmark_repo_dir (i.e. alongside the `repo/` and `data/`
directories) as `paper.pdf`. The paper helps downstream auditors understand the
benchmark's methodology and evaluation design.

### Step 1c: Populate the data directory (if needed)
If `benchmark_data_dir` is empty or does not exist, you MUST populate it with
the benchmark's evaluation dataset.

Read the cloned repo's README, paper references, and source code to understand
how this benchmark's data is structured and where it comes from. Then acquire
the data using whatever method is appropriate - download it, copy it from the
repo, extract it from source code, or generate it using the repo's own scripts.

The goal is to have the actual evaluation tasks/problems available as files in
`benchmark_data_dir` so the collector script in Phase 2 can parse them.

- If acquisition requires authentication, load credentials from a `.env` file
  at the working directory (e.g., HF_TOKEN, KAGGLE_KEY) before downloading.
- Extract any downloaded archives (zip/tar/gz) into `benchmark_data_dir`,
  organize the contents so each task's files live under a self-contained
  per-task location, and ensure each TaskConfig path (task_bundle_path,
  tests_ref, solution_ref) resolves to a real file.

If you cannot obtain the data after investigating the repo, document what you
tried and proceed - but note "insufficient_data" in any output.

### Step 1d: Benchmark-specific data acquisition (REQUIRED, when supplied)
The following instructions override the generic Step 1c guidance for
this benchmark. Follow them exactly to populate `benchmark_data_dir`:

[DATA_ACQUISITION_HINT]

## Phase 2: Collect evidence

The goal of this phase is to produce a structured inventory of every task in the
benchmark, with clear pointers back to the data you acquired in Phase 1.
The downstream per-task audit agent will rely on these pointers to inspect
individual tasks, their evaluation criteria, and their test/scoring logic.

1. Read the dataclass definitions in [MODELS_PATH] to understand ArtifactManifest,
   TaskEntry, and TaskConfig schemas.

2. Explore the benchmark data directory (and repo if available) to understand:
   - What constitutes a single "task" or "problem" in this benchmark
   - What the evaluation input (prompt, question, scenario) looks like
   - What the expected/reference answer or success criteria is
   - How tasks are evaluated - what scripts, test harnesses, or scoring
     functions determine correctness

3. Write a Python 3 collector script to: [COLLECTOR_PATH]
   - Accept --manifest-path and --task-config-dir CLI arguments
   - Load and parse the benchmark dataset from benchmark_data_dir
   - Discover all tasks/problems in the dataset
   - For each task, extract the fields described below
   - Write a YAML ArtifactManifest to --manifest-path
   - Write per-task TaskConfig JSON files to --task-config-dir/<task_id>/task_config.json
   - Do NOT write EvalConfig files (there are no agent eval runs for this benchmark)

4. Run the collector script:
   python [COLLECTOR_PATH] --manifest-path [MANIFEST_PATH] --task-config-dir [TASK_CONFIG_DIR]

You are responsible for both writing AND executing the collector.
The runtime will only validate the outputs afterward.

Benchmark inputs:
- benchmark_name: [BENCHMARK_NAME]
- benchmark_type: [BENCHMARK_TYPE]
- job_type: [JOB_TYPE]
- benchmark_data_dir: [BENCHMARK_DATA_DIR]
- benchmark_repo_dir: [BENCHMARK_REPO_DIR]
- code_url: [CODE_URL]
- output_root: [OUTPUT_ROOT]

### TaskConfig field rules

These are the critical fields for downstream task auditing. The per-task audit
agent uses `task_bundle_path`, `tests_ref`, and `solution_ref` to navigate
directly to the relevant files - populate them carefully.

Identity & status (static benchmark - no eval runs):
- status: "unscored"
- n_evals: 0, n_passed: 0, n_failed: 0
- primary_eval_id: null

Task content - link these back to the actual files from Phase 1:
- task_bundle_path: path to the root directory or primary data file for this
  task (e.g. the task's subdirectory in benchmark_data_dir, or the source
  JSON/JSONL file that contains it). This is the entry point for the auditor.
- instruction_text: the full problem/question text for this task (inline it)
- instruction_path: path to the file containing the instruction (if it exists
  as a standalone file), null otherwise
- reference_answer: the gold/expected outcome for this task (inline the text).
  For standard benchmarks this is the correct answer or solution. For
  adversarial/security benchmarks, describe both what a successful attack
  looks like AND what a secure/correct agent response should be - make the
  ground-truth label clear (e.g. "attack succeeds" vs "agent resists"). For
  tasks with complex answers (code, structured output, multi-step reasoning),
  include enough to understand what correct behavior looks like and how it
  is scored.
- solution_ref: path to the reference answer or gold solution file on disk
  (null if the answer is only available inline or not on disk)
- tests_ref: path to the evaluation/test script, scoring function, or test
  harness that determines whether a response to this task is correct. Look in
  the repo for eval scripts, pytest files, grading functions, or judge prompts.
  This is critical - the task auditor needs to understand HOW correctness is
  determined, not just WHAT the answer is. Set to null only if no evaluation
  logic is discoverable.
- audit_notes: free-text guidance for the downstream task audit agent. Write
  this as a handoff note summarizing what you learned while exploring the
  benchmark. Include: how this specific task maps to source data (e.g. which
  file, array index, or key identifies it), where the eval logic for this
  task's eval_type lives in the codebase (file paths and class/function names),
  any benchmark-specific context the auditor needs (e.g. how scoring works,
  what the eval criteria actually check, which parameters are template-resolved
  at runtime). The auditor will read the actual files - your job is to tell
  them WHERE to look and WHAT to look for.
  If the task references external artifacts (images, audio, video, PDFs, or
  any other files the problem statement or reference answer depends on),
  resolve each artifact to its actual on-disk location and list the absolute
  paths under a "Referenced artifacts:" section. The auditor is required to
  read every path in audit_notes, so this is how you make those artifacts
  part of the audit. Do not trust paths stored inside upstream data files
  without verifying they exist - re-resolve against benchmark_data_dir.

ArtifactManifest rules:
- Each TaskEntry must have a task_id and an empty eval_ids list
- source_data_dir: set to benchmark_data_dir

Conventions:
- Emit absolute paths everywhere
- Use yaml.safe_dump(..., sort_keys=False) for the manifest
- Use json.dumps(..., indent=2, sort_keys=True) for TaskConfig
- Prefer deterministic ordering by task_id
- For missing optional fields, use null
- If the dataset is large (>5000 tasks), include all tasks - the downstream
  audit step handles sampling
\end{tcblisting}

\subsection{Audit prompts}
\label{app:prompts}

The auditor agent receives a single rendered prompt per task. Two
templates exist: a \emph{trajectory-mode} template
(Listing~\ref{lst:prompt:trajectory}) used when an agent run is
available, and a \emph{static-mode} template
(Listing~\ref{lst:prompt:static}) used when only the task definition
is available. Both inline the audit-time rubric verbatim
(Listing~\ref{lst:rubric:audit}); the rubric is the same
single-dimension prompt-vs-tests instrument referenced in
\S\ref{sec:framework:auditing_protocol}. Tokens in
square brackets (e.g.\ \texttt{[TASK\_ID]},
\texttt{[INSTRUCTION\_TEXT]}, \texttt{[RUBRIC\_TEXT]}) are
substituted at render time from the task configuration and the
evidence-collector manifest; the rest of the prompt is fixed.

\begin{tcblisting}{
  enhanced, breakable,
  colback=gray!3, colframe=black!55,
  title={Listing 2: Trajectory-mode audit prompt template},
  fonttitle=\bfseries, coltitle=white, colbacktitle=black!75,
  boxrule=0.4pt, arc=2pt,
  left=6pt, right=6pt, top=4pt, bottom=4pt,
  listing only,
  label={lst:prompt:trajectory},
  listing options={
    basicstyle=\ttfamily\scriptsize,
    breaklines=true,
    breakatwhitespace=true,
    columns=fullflexible,
    keepspaces=true,
    showstringspaces=false,
    upquote=true,
  },
}
You are auditing one benchmark task to determine whether failures (or even passes) reveal issues in the benchmark itself - not in the agent.

Your goal: decide whether the task's prompt, environment, and test suite are fair and well-specified. If they are, report no findings. If not, document each issue with evidence.

## Workflow
1. Read the rubric below. It defines what counts as a finding, how to distinguish agent error from genuine task issues, and the severity scale. Use its examples and criteria to guide your judgment.
2. Explore the eval artifacts further below. Open and read every path provided. Check them against the rubric and the prompt. Do not reason about file contents without reading them first.
3. Ground every claim in what you observed. Cite concrete file paths as evidence. If the evidence does not support a benchmark issue, return an empty findings list.

Return exactly one JSON object and nothing else.

Important contract:
- Artifacts are passed as filesystem paths, not inline content. You must open and read those files before making claims.
- Distinguish prompt/specification issues from agent issues.
- A finding is a benchmark issue, not an agent mistake. If the agent simply wrote wrong code and the task is fair, that is severity 0 with no findings.

## Per-task rubric
[RUBRIC_TEXT - see Listing 4]

## Evals artifacts field guide
- instruction_text: the exact prompt the agent received - this is what you audit for clarity and completeness. Check this against the reference solution, tests and agent evals to see if the prompt is underspecified or ambiguous.

Per-eval artifact paths (inside each selected_evals entry):
- metrics_path: structured test metrics - pass/fail counts and per-test details.
- test_output_path: raw stdout from the test harness run. Read this to see which tests passed/failed and why (assertion errors, tracebacks, etc.).
- prediction_path: the agent's final output (patch, code, or text).
- trajectory_path: the agent's full conversation/action log (JSON) - shows reasoning, tool calls, and decisions made during the attempt. Read this to understand why the agent succeeded or failed and whether confusion stemmed from benchmark issues.

Task-level paths (present when available):
- tests_ref: the test suite used to evaluate solutions - may be a directory of test files or a single eval script. Read these to check if tests are fair, narrow, or misaligned with the prompt.
- environment_ref: files placed in the container before the agent starts (starter code, data files, configs). Check for mismatches with what the prompt describes.
- solution_ref: the reference/gold solution. Compare against agent output to understand what was expected.
- task_bundle_path: JSON file containing the task definition (e.g. SWE-bench problem_statement, repo, patch, test_patch). Shown when tests_ref/environment_ref/solution_ref are not all available.

## Evals artifacts
{
  "task_id": "[TASK_ID]",
  "instruction_text": "[INSTRUCTION_TEXT]",
  "selected_evals": [
    {
      "eval_id": "[EVAL_ID]",
      "status": "[passed|failed]",
      "metrics_path": "[METRICS_PATH]",
      "test_output_path": "[TEST_OUTPUT_PATH]",
      "prediction_path": "[PREDICTION_PATH]",
      "trajectory_path": "[TRAJECTORY_PATH]"
    }
  ],
  "tests_ref": "[TESTS_REF_PATH]",
  "environment_ref": "[ENVIRONMENT_REF_PATH]",
  "solution_ref": "[SOLUTION_REF_PATH]",
  "audit_notes": "[AUDIT_NOTES]"
}
\end{tcblisting}

\begin{tcblisting}{
  enhanced, breakable,
  colback=gray!3, colframe=black!55,
  title={Listing 3: Static-mode audit prompt template},
  fonttitle=\bfseries, coltitle=white, colbacktitle=black!75,
  boxrule=0.4pt, arc=2pt,
  left=6pt, right=6pt, top=4pt, bottom=4pt,
  listing only,
  label={lst:prompt:static},
  listing options={
    basicstyle=\ttfamily\scriptsize,
    breaklines=true,
    breakatwhitespace=true,
    columns=fullflexible,
    keepspaces=true,
    showstringspaces=false,
    upquote=true,
  },
}
You are auditing one benchmark task WITHOUT execution artifacts - no agent trajectories, test outputs, or eval run data. Your goal: determine whether the task's problem statement, reference answer, and evaluation logic are well-specified and fair.

If they are, report no findings. If not, document each issue with evidence.

## Workflow
1. Read the rubric below. It defines what counts as a finding, the severity scale, and the distinction between benchmark issues and expected difficulty.
2. Read the problem statement and reference answer provided below.
3. If a benchmark repository or eval code is available (paths below), open and inspect the evaluation scripts, test harness, and scoring logic for fairness and correctness.
4. Ground every claim in what you observed. Cite concrete file paths or inline content as evidence. If the evidence does not support a benchmark issue, return an empty findings list.

Return exactly one JSON object and nothing else.

Important contract:
- You are auditing the benchmark task itself, not any model's performance.
- A finding is an issue in the task's definition, reference answer, or evaluation logic - not normal difficulty or expected domain knowledge.
- If `audit_notes` contains any filesystem paths (e.g. reference inputs, gold deliverables, rubric files, eval scripts), those paths are a critical part of the audit. You MUST open and read through every such path before forming conclusions. Do not reason about their contents based on filenames, directory listings, or the notes alone - read the files. This applies to PDFs, ZIPs, code directories, and any other artifact referenced by path.

## Per-task rubric
[RUBRIC_TEXT - see Listing 4]

## Artifacts field guide
- instruction_text: the exact problem/question text - audit this for clarity and completeness.
- reference_answer: the gold/reference answer - check this for correctness, completeness, and whether alternative valid answers exist.

Task-level paths (present when available):
- solution_ref: path to reference solution file(s). Compare against reference_answer to check consistency.
- tests_ref: path to test/eval scripts. Inspect for fairness, narrowness, or misalignment with the problem statement.
- benchmark_repo_dir: root of the benchmark repository. Explore eval code, scoring scripts, and data loading logic.
- benchmark_data_dir: root of the dataset. Inspect data files if needed to verify task content.
- task_bundle_path: path to the source data file or directory for this task (e.g. the original JSON entry, task subdirectory). When solution_ref is null, this is the best place to find the raw task definition including structured eval criteria, scoring parameters, and the original reference answer.
- audit_notes: guidance from the evidence collector about how to navigate this task's data - where the eval logic lives, how the task maps to source data, and benchmark-specific context. Use this as a starting point, but verify claims by reading the actual files. If the notes list filesystem paths (reference inputs, gold deliverables, rubric files, etc.), treat those as required reading - open and read every one before drawing conclusions.

## Task artifacts
{
  "task_id": "[TASK_ID]",
  "instruction_text": "[INSTRUCTION_TEXT]",
  "reference_answer": "[REFERENCE_ANSWER]",
  "solution_ref": "[SOLUTION_REF_PATH]",
  "tests_ref": "[TESTS_REF_PATH]",
  "task_bundle_path": "[TASK_BUNDLE_PATH]",
  "audit_notes": "[AUDIT_NOTES]",
  "benchmark_repo_dir": "[BENCHMARK_REPO_DIR]",
  "benchmark_data_dir": "[BENCHMARK_DATA_DIR]"
}
\end{tcblisting}

\begin{tcblisting}{
  enhanced, breakable,
  colback=gray!3, colframe=black!55,
  title={Listing 4: Audit-time per-task rubric (substituted into [RUBRIC\_TEXT])},
  fonttitle=\bfseries, coltitle=white, colbacktitle=black!75,
  boxrule=0.4pt, arc=2pt,
  left=6pt, right=6pt, top=4pt, bottom=4pt,
  listing only,
  label={lst:rubric:audit},
  listing options={
    basicstyle=\ttfamily\scriptsize,
    breaklines=true,
    breakatwhitespace=true,
    columns=fullflexible,
    keepspaces=true,
    showstringspaces=false,
    upquote=true,
  },
}
Per-task audit rubric for benchmarks. A finding is a gap between what the prompt tells the agent and what the tests actually require.

Findings fall into one of three categories: ambiguity_underspecification (the prompt is unclear, vague, or missing information needed to pass), tests (the tests are too narrow, too broad, or check something other than what the prompt describes), or environment_conflict (the provided environment, files, or dependencies contradict or block what the prompt asks for).

The agent cannot see the tests. Every piece of information needed to pass must be discoverable from the prompt, the provided files, or standard domain knowledge alone. If it is not, that is a task issue - regardless of whether the agent's implementation was also flawed.

Distinguishing agent error from task issues:
  Before assigning any finding, ask: could a competent developer, reading only the prompt, reliably produce a passing solution?

  This is agent error (not a finding):
  - The prompt clearly states the requirement but the agent misreads, ignores, or forgets it.
  - The agent makes an implementation mistake unrelated to prompt clarity.
  - The agent runs out of time due to an inefficient strategy, not because the task is infeasible.
  - The agent fails to use a standard technique the prompt reasonably assumes as domain knowledge.

  This is a task issue (a finding):
  - The information needed to pass the test is not present in the prompt, the provided files, or standard domain knowledge. The agent would have to guess or reverse-engineer the test.
  - Two competent developers could reasonably disagree on what the prompt asks, and the tests only accept one interpretation.
  - The prompt actively misleads: an example or instruction points toward a solution the tests reject.
  - The tests enforce specific implementation details - function names, internal structure, module organization - that the prompt never specifies. A correct solution using different internals would fail.
  - The test rejects a solution that correctly implements what the prompt describes.

  Important: an incomplete implementation is not automatically agent error. Before reaching that conclusion, check whether the missing pieces are things the prompt told the agent to build, or things only the tests require. If the tests demand specific internal structure that the prompt never specifies, the incompleteness reflects a task issue, not a skill gap.

  Litmus test: give the prompt to five competent developers in the relevant domain, without showing them the tests. If most would produce solutions the tests reject, the task has a finding.

Examples of hard judgment calls:
  - The prompt references a live external resource at a historical point but provides no frozen snapshot. If the reference solution reveals a discoverable path via standard tools (git history, well-known APIs), severity is 1. If no standard path exists, severity is 2.
  - The test silently replaces input data with a different version before evaluation, and the prompt gives no indication. This is a finding even if the agent's solution was also wrong for other reasons.
  - The tests import specific helper functions by name from the reference solution's module. The agent would need to create those exact functions in those exact locations. This is a finding - the prompt describes the feature, not the internal API.
  - The prompt is vague but the expected approach is domain-standard. A developer can infer it no problem. This is severity 0, not severity 1 - the gap exists, it is just small.

Severity scale and calibration:
  Before assigning severity, cross-check the task instruction prompt against the reference solution and tests. Read the reference solution to identify its approach. Read the test suite and the prompt to check whether the prompt is underspecified or ambiguous, and whether the tests are too narrow (only accepting one approach) or too broad (accepting trivially wrong solutions). If the prompt leaves room for multiple reasonable interpretations and the test suite enforces just one, that gap is a finding - even if the reference solution makes the intended interpretation obvious. Then ask: could a competent developer, reading only the prompt, reasonably discover the expected approach using standard tools and practices? Use that answer to assign severity:

  0 (None): The prompt is clear and sufficient. The tests fairly evaluate what the prompt asks for. Any failure is attributable to the agent's own mistakes or skill gaps.
  1 (Minor): A discoverable gap. Reading only the prompt, a competent developer in the relevant domain can reach the expected solution - but not immediately. The bridge requires deliberate work: consulting git history, a known API, reference docs, or reasoning through which of a few standard approaches the tests likely want. Some developers will cross it, others will pick a different reasonable interpretation that the tests reject.
  2 (Major): It is almost impossible to understand what you are being asked to do without further information. The prompt omits or obscures key details that no amount of domain expertise can compensate for. The tests are too narrow, too broad, or look for something different from what the task description asks. Information may be missing, contradictory, or actively misleading.

  Include this calibration reasoning in each finding's evidence.
\end{tcblisting}

\subsection{Auditor reproducibility}
\label{app:reproducibility}

\paragraph{Claude Code configuration.}
We pin the auditor harness to the same Claude Code CLI
configuration described in \S\ref{sec:audit:experiment_setup:configuration}. Every
audit in this paper was produced by the following stack:

\begin{itemize}
  \item \textbf{CLI}: Claude Code, version
    \texttt{v2.1.96} (\texttt{claude --version}; pinned for the
    duration of the study).
  \item \textbf{Model}: Anthropic Claude
    \texttt{claude-opus-4-7}, invoked via
    \texttt{claude --model opus} (alias resolves to Opus~4.7 at
    the pinned CLI version). Sampling, system prompt, and agent
    loop are left at the CLI defaults; we do not override
    temperature or max-tokens.
  \item \textbf{Tools}: the default Claude Code tool set --
    file read/write (\texttt{Read}, \texttt{Write},
    \texttt{Edit}), shell (\texttt{Bash}), and search
    (\texttt{Grep}, \texttt{Glob}). We do not enable any MCP
    servers or custom tools, and we do not pass
    \texttt{--allowed-tools} or \texttt{--disallowed-tools}.
  \item \textbf{Permissions}: \texttt{--permission-mode
    bypassPermissions} so the harness runs non-interactively;
    audits run inside a sandbox that restricts the agent to the
    per-task evidence directory plus the read-only benchmark
    repository checkout.
  \item \textbf{Invocation mode}: headless / one-shot, via
    \texttt{claude -p <prompt> --output-format json}; each task
    is a fresh session (no \texttt{--resume}, no \texttt{--continue})
    so audits are independent and order-insensitive.
\end{itemize}

The exact prompts substituted into each session are reproduced
verbatim in Listings~\ref{lst:prompt:trajectory}--\ref{lst:rubric:audit}
above; the evidence-collector manifests and run scripts that
assemble the per-task artifact bundles will be released with the
code.

% !TEX root = ../neurips_2026.tex
% Per-task breakdown of Terminal-Bench~2 PR #53 issues and how
% the trajectory audit aligned with each one. Source data:
% bench_audit/benchmarks/leaderboards/terminal-bench_v2/pr53_changes/
%   issues_per_task.json
% Audit run: tb2/tb2__ambiguity_v3_claude_opus_4_7__actual.

\begin{table}[t]
  \centering
  \scriptsize
  \setlength{\tabcolsep}{4pt}
  \renewcommand{\arraystretch}{1.05}
  \caption{Per-task breakdown of Terminal-Bench~2 PR\,\#53. Each
  row corresponds to one PR-flagged issue, with the rubric
  category from the source JSON's \texttt{issue\_category} field
  (INS = Instruction, ENV = Environment, EVA = Evaluation) and a
  one-line summary. The ``Audit'' column records whether our
  trajectory audit produced an \emph{aligned} finding (same
  root cause as the PR fix), a \emph{partial} match (same
  phenomenon observed but framed differently or without a
  formal finding being filed), or \emph{none} (no finding
  addressed this issue). Four PR-touched tasks have no
  in-rubric issue and are listed with the maintainer's reason
  for the patch (resource bump, oracle-only fix, etc.); these
  sit outside the task-level rubric and are not counted in the
  recall.}
  \label{tab:tb2_pr53_per_task}
  \begin{tabular}{@{}p{1.45in} c p{2.95in} c@{}}
    \toprule
    \textbf{Task ID} & \textbf{Cat.} & \textbf{Summary of issue} & \textbf{Audit} \\
    \midrule
    \texttt{adaptive-rejection-sampler} & INS & Function signature unspecified in prompt. & Aligned \\
    \texttt{build-pmars}                & EVA & Tests hardcode a Debian source-tree layout the prompt does not require. & Partial \\
    \texttt{caffe-cifar-10}             & EVA & Solver config path ambiguous between prompt and test. & Aligned \\
    \texttt{caffe-cifar-10}             & EVA & Test hardcoded to \texttt{make}, rejecting valid \texttt{cmake} Caffe builds. & None \\
    \texttt{compile-compcert}           & --- & No in-rubric issue: resource bump (\texttt{nproc} wrapper to avoid host-CPU OOMs). & --- \\
    \texttt{configure-git-webserver}    & INS & Prompt vs.\ test contract mismatch on auth (prompt promised auth was handled; grader required password SSH). & Aligned \\
    \texttt{extract-moves-from-video}   & ENV & Live YouTube URL is an unreliable external dependency (PR bundles the video into the image). & Aligned \\
    \texttt{filter-js-from-html}        & INS & Prompt vs.\ test contradiction on HTML formatting (prompt forbids reformatting; tests use BeautifulSoup normalization). & Aligned \\
    \texttt{filter-js-from-html}        & ENV & Tests fetch XSS vectors from GitHub at test time (network dependency; not addressed by PR). & Aligned \\
    \texttt{fix-git}                    & ENV & Installation patch files left in the container enabled cheating; grader switched to hardcoded MD5 hashes. & None \\
    \texttt{hf-model-inference}         & EVA & Test assumed \texttt{save\_pretrained} layout only; HuggingFace cache layout silently rejected. & Partial \\
    \texttt{install-windows-3.11}       & INS & QEMU monitor socket path hardcoded in test but unspecified in prompt. & Aligned \\
    \texttt{make-doom-for-mips}         & --- & No in-rubric issue: oracle-only fix (prepend \texttt{apt-get update} to avoid stale-index 404s). & --- \\
    \texttt{mteb-leaderboard}           & INS & Prompt did not state the all-28-task coverage requirement the grader filters on. & Aligned \\
    \texttt{mteb-leaderboard}           & INS & Canonical benchmark-version string \texttt{MTEB(Scandinavian, v1)} not named in prompt. & Aligned \\
    \texttt{mteb-retrieve}              & INS & Prompt allowed bypassing the \texttt{mteb} encoding pipeline (skipping asymmetric \texttt{PromptType} prefixes). & Aligned \\
    \texttt{polyglot-c-py}              & EVA & Test enforced unstated ``source-file only'' constraint via strict \texttt{os.listdir} equality. & Aligned \\
    \texttt{polyglot-rust-c}            & EVA & Same class as \texttt{polyglot-c-py}: strict \texttt{os.listdir} equality on the source directory. & Aligned \\
    \texttt{protein-assembly}           & ENV & PDB data drift: chromophore residues now return \texttt{X} in FASTA, breaking the oracle's codon optimisation. & Partial \\
    \texttt{query-optimize}             & INS & Prompt did not forbid DB modification, but grader checks SHA256 of the database file. & None \\
    \texttt{rstan-to-pystan}            & --- & No in-rubric issue: oracle-only fix (use \texttt{/opt/py310} venv; pin \texttt{setuptools<71}). & --- \\
    \texttt{sam-cell-seg}               & INS & Prompt vs.\ test argument-form mismatch (positional names in prompt; required \texttt{--}options in test). & Aligned \\
    \texttt{torch-pipeline-parallelism} & --- & No in-rubric issue: resource bump (memory $4{\rightarrow}8$\,GB to avoid wheel-extraction OOMs). & --- \\
    \texttt{torch-tensor-parallelism}   & INS & \texttt{RowParallelLinear.forward()} input contract unstated; agents that scatter internally double-scatter. & Aligned \\
    \texttt{train-fasttext}             & EVA & Verifier used \texttt{fasttext-wheel} Python bindings that fail to load models the same library produced. & None \\
    \bottomrule
  \end{tabular}
\end{table}

\subsection{Auditor cost}
\label{app:cost}

To gauge the dollar cost of running our auditor, we sampled
10 task audits uniformly at random from each of six
benchmark runs and computed per-conversation cost from the
Claude Code result-line token usage (\texttt{input\_tokens},
\texttt{output\_tokens}, \texttt{cache\_creation\_input\_tokens},
\texttt{cache\_read\_input\_tokens}) using the published
Anthropic API price list\footnote{\url{https://platform.claude.com/docs/en/about-claude/pricing} (accessed May 2026).}
for Claude Opus~4.7: \$5/MTok base input,
\$25/MTok output, \$6.25/MTok 5-minute cache write,
\$10/MTok 1-hour cache write, and \$0.50/MTok cache read.
Across the six samples, the mean cost per static task audit was
\$0.42 on SWE-bench Verified ($n{=}500$),
\$0.73 on FinanceAgent ($n{=}50$),
\$0.40 on MMMU-Pro ($n{=}503$),
\$0.64 on Terminal-Bench~2 ($n{=}89$),
\$0.80 on OSWorld-Verified ($n{=}369$), and
\$0.61 on Humanity's Last Exam ($n{=}498$),
for an unweighted average of roughly \$0.60 per task audit.

Two factors dominate this cost. First, prompt caching is
heavily exploited: across the sampled conversations the
average audit consumes on the order of $1$--$6{\times}10^{5}$
cached input tokens at the 10\% read rate, while only a few
tens of thousands of tokens are paid at the higher
cache-write rate; this is what keeps a multi-turn agentic
audit at sub-dollar cost. Second, output token volume and
turn count track task complexity --- OSWorld-Verified and
Terminal-Bench~2 audits average $10$--$15$ turns and
$\sim 10$k output tokens, while MMMU-Pro audits finish in
${\sim}6$ turns, explaining most of the spread across
benchmarks. At these unit costs, auditing an entire
500-task benchmark with our default static-context
configuration costs on the order of \$200--\$400, well
within the budget of any team that can afford to run the
underlying benchmark itself.

\section{Validation details}
\label{app:validation}

This appendix backs the three validation strategies in
\S\ref{sec:audit:validation} with the per-task lists and protocol
details that did not fit in the body.

\subsection{Terminal-Bench~2 PR\,\#53 overlap details}
\label{app:tb2_pr53}

This subsection breaks down the Terminal-Bench~2 PR\,\#53 overlap
result. The task-by-task issue list is in Table~\ref{tab:tb2_pr53_per_task}, which lists each PR-flagged
issue, a one-line summary, and whether our trajectory audit
produced an aligned, partial, or no-finding match. PR\,\#53 addresses 22 tasks in its body. Going through each
task's PR diff, we extract 21 individual findings in
total (Table~\ref{tab:tb2_pr53_per_task}): each finding
corresponds to one issue fixed by the PR that could plausibly
confuse an agent or LLM (e.g., a prompt--test contract
mismatch, an unstated interface convention, or a brittle
external dependency). We exclude PR changes that are
due-diligence patches rather than agent-facing defects --
resource bumps, oracle-only fixes, dependency rot, and minor
cleanups -- which together account for the four PR-touched
tasks with no extracted finding (\texttt{compile-compcert},
\texttt{make-doom-for-mips}, \texttt{rstan-to-pystan},
\texttt{torch-pipeline-parallelism}). 

Against this 21-finding pool, our trajectory audit on Terminal-Bench~2
strictly matches (same root cause as the PR fix) 14
findings and aligned-or-partial matches (same
phenomenon observed, possibly framed differently or without a
formal finding filed) 17 findings, giving strict recall
\(14/21=66.7\%\) and aligned-or-partial recall
\(17/21=81.0\%\).

\paragraph{(a) Audit findings on tasks PR\,\#53 does not touch.}
The trajectory audit flagged 11 additional Terminal-Bench~2
tasks not addressed by PR\,\#53 at all. Seven of these carry a
Major-severity finding. For those seven tasks, we additionally
reviewed each finding by hand and record the review outcome in
the ``Audit'' column of \Cref{tab:tb2_pr53_audit_only}. Across
the 11 manually reviewed findings, 8 are strictly aligned and one
is partially aligned, giving strict precision \(8/11=72.7\%\) and
aligned-or-partial precision \(9/11=81.8\%\). These ABA-only
findings show that the PR-relative precision reported in the main
text is conservative: PR\,\#53 covers a specific maintainer patch
set, while ABA surfaces additional valid issues outside that scope,
complementing human review by expanding coverage without the
accuracy loss that would make the extra findings unusable.

\begin{table}[t]
  \centering
  \scriptsize
  \setlength{\tabcolsep}{3pt}
  \caption{Audit tasks flagged on Terminal-Bench~2 outside the
  scope of PR\,\#53. For the seven Major tasks reviewed by hand,
  unindented rows identify the task and indented rows identify
  each finding, with issue-level severity, category, and human
  review outcome. Categories are INS = Instruction, ENV =
  Environment, EVA = Evaluation. ``--'' indicates that the row was
  not part of this manual review.}
  \label{tab:tb2_pr53_audit_only}
  \begin{tabular}{@{}p{2.65in}ccp{0.62in}@{}}
    \toprule
    \textbf{Task / finding} &
      \textbf{Issue sev.} &
      \textbf{Category} &
      \textbf{Audit} \\
    \midrule
    \texttt{count-dataset-tokens} & & & \\
    \hspace{0.8em}\emph{Issue~1}: tokenization aggregation ambiguity
      & Major & INS & Aligned \\
    \hspace{0.8em}\emph{Issue~2}: science-domain mapping overclaimed
      & Minor & INS & Partial \\
    \texttt{db-wal-recovery} & & & \\
    \hspace{0.8em}\emph{Issue~1}: destructive WAL access
      & Major & ENV & Aligned \\
    \texttt{largest-eigenval} & & & \\
    \hspace{0.8em}\emph{Issue~1}: microsecond speed gate
      & Major & EVA & Aligned \\
    \hspace{0.8em}\emph{Issue~2}: public helper omits odd sizes
      & Minor & INS & Aligned \\
    \texttt{model-extraction-relu-logits} & & & \\
    \hspace{0.8em}\emph{Issue~1}: hidden RNG side effect
      & Major & ENV & Aligned \\
    \texttt{pytorch-model-cli} & & & \\
    \hspace{0.8em}\emph{Issue~1}: unspecified image preprocessing
      & Major & INS & Aligned \\
    \texttt{raman-fitting} & & & \\
    \hspace{0.8em}\emph{Issue~1}: peak model and parameterization
      & Major & INS & Aligned \\
    \hspace{0.8em}\emph{Issue~2}: input format and x-axis units
      & Major & INS & Aligned \\
    \hspace{0.8em}\emph{Issue~3}: narrow parameter tests
      & Minor & EVA & None \\
    \texttt{video-processing} & & & \\
    \hspace{0.8em}\emph{Issue~1}: takeoff/landing frame definition
      & Major & INS & None \\
    \texttt{gpt2-codegolf}             & Minor & ENV       & -- \\
    \texttt{mailman}                   & Minor & EVA       & -- \\
    \texttt{make-mips-interpreter}     & Minor & EVA       & -- \\
    \texttt{qemu-alpine-ssh}           & Minor & EVA       & -- \\
    \bottomrule
  \end{tabular}
\end{table}

\paragraph{(b) PR\,\#53 tasks without a matching audit finding.}
Four PR\,\#53 tasks have no trajectory-audit finding addressing any of their PR-extracted findings -- neither a strict nor an aligned-or-partial match. They are \texttt{compile-compcert},
\texttt{make-doom-for-mips}, \texttt{rstan-to-pystan},
\texttt{torch-pipeline-parallelism}.

\subsection{BenchGuard match criteria}
\label{app:benchguard}

To make the BixBench and SAB rows of Table~\ref{tab:audit:benchguard}
strictly comparable to \citet{tu2026benchguard}, we adopt the
BenchGuard evaluation protocol verbatim: same gold-issue sets, same
alignment-judge prompt, same scoring scripts, and same judge model
(\texttt{gemini-3-flash-preview} at temperature~$0$). Concretely,
we feed our ABA findings on the SAB and BIXBench Verified-50 task
sets into BenchGuard's released evaluation pipeline (\texttt{eval/normalize.py}
$\rightarrow$ \texttt{eval/match.py} $\rightarrow$ \texttt{eval/metrics.py})
without modification or any prior candidate pruning. For each
$(\text{gold issue},\,\text{finding})$ pair the alignment judge
emits one of \textsc{aligned} (a single fix would resolve both),
\textsc{partial} (correct functional area but a genuinely different
problem), or \textsc{unrelated}. A gold issue is then counted as
recovered at the strict bar (s) if any finding is judged
\textsc{aligned} against it, and at the partial bar (p) if any
finding is at least \textsc{partial}; precision is computed
analogously over findings on the revised-task subset, taking each
finding's best verdict across gold issues. Strict and partial
recall and precision in Table~\ref{tab:audit:benchguard} are
exactly BenchGuard's $\mathrm{Recall}_A$/$\mathrm{Recall}_{A+P}$
and $\mathrm{Precision}_A$/$\mathrm{Precision}_{A+P}$. We refer the
reader to \citet{tu2026benchguard}, in particular their Section~4.2
and Appendix~J, for the full alignment-judge prompt, gold-issue
construction, and ensemble metrics.

\subsection{Benchmarks covered by manual confirmation}
\label{app:validation:manual_benchmarks}

The 54 Major and 30 Minor static-mode findings reviewed in
\S\ref{sec:audit:validation:manual} are sampled across the
following 12 frontier benchmarks, which together span coding,
agentic tool use, computer use, browsing, expert reasoning,
mathematics, multimodal understanding, and economically valuable
professional work: Finance Agent~v1.1~\citep{bigeard2025finance},
GDPval~\citep{openai2025gdpval}, Humanity's Last
Exam~\citep{phan2025lastexam},
IMO-AnswerBench~\citep{luong-etal-2025-towards},
MMMU-Pro~\citep{yue2024mmmupro},
OSWorld-Verified~\citep{xie2024osworld,xlang2025osworldverified},
$\tau^2$-Bench (Telecom)~\citep{yao2025taubench,barres2025tau2},
Toolathlon~\citep{li2025toolathlon},
SWE-bench Multilingual~\citep{jimenez2024swebench},
SWE-bench Verified~\citep{openai2024swebenchverified},
Terminal-Bench~2~\citep{merrill2026terminalbench}, and
BrowseComp~\citep{openai2025browsecomp}. The trajectory-mode set of
25 Major findings is sampled from SWE-bench
Verified~\citep{openai2024swebenchverified}.

\section{Leaderboard sensitivity}
\label{app:leaderboards}

This section backs the leaderboard-sensitivity claims in
\S\ref{sec:ablations} with the per-benchmark figure, and a check that the trajectory auditor's
extra findings are not driven by agent failure. For each benchmark
with a wired leaderboard, we record the snapshot source (date and
URL or repository), the submitted-systems list, and the exact pre-
and post-exclusion pass rates per system at severity thresholds
$s\in\{\geq 1,\geq 2,=1\}$, together with the top-$k$ rankings on
either side of the exclusion. 

\subsection{Per-model score change after excluding flagged tasks}
\label{app:leaderboards:meandelta}

\begin{figure}[t!]
  \centering
  \includegraphics[width=\linewidth]{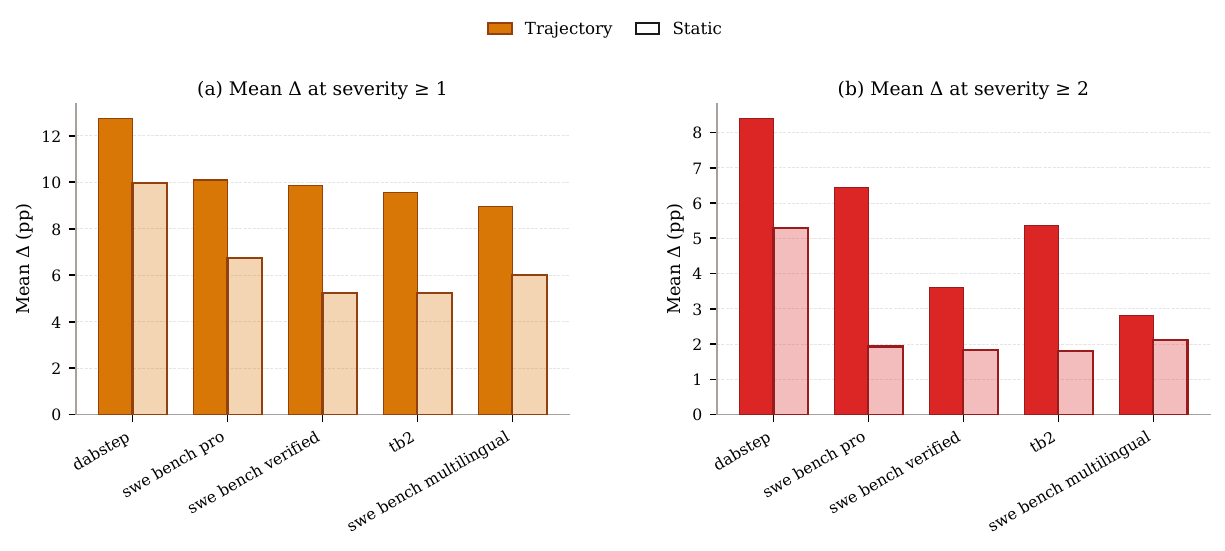}
  \caption{Mean per-model score change (in percentage points) after
  excluding audit-flagged tasks, on the five benchmarks for which
  per-instance leaderboard data are available. (a) removes tasks
  flagged at Minor or Major severity (severity~$\geq 1$); (b) removes
  only tasks flagged Major (severity~$\geq 2$). Bars average over
  all models reported on each benchmark.}
  \label{fig:ablation:leaderboard_mean_delta}
\end{figure}

We select the five audited benchmarks for which per-instance
leaderboard scores are publicly available -- {\tt dabstep},
{\tt swe bench pro}, {\tt swe bench verified}, {\tt tb2}, and
{\tt swe bench multilingual} -- and compute, for each model on
each benchmark, the change in its reported pass rate when
audit-flagged tasks are removed from the evaluation set.
Figure~\ref{fig:ablation:leaderboard_mean_delta} shows the mean
change across models under two exclusion rules: panel~(a) drops
tasks flagged at Minor or Major severity, and panel~(b) drops only
Major-flagged tasks.

Across all five benchmarks, removing flagged tasks raises the
average model score, and the broader exclusion rule in~(a)
produces a uniformly larger lift than the Major-only rule in~(b).
The effect is not marginal: under~(a) the mean per-model gain
reaches roughly 13~percentage points on {\tt dabstep} and remains
in the multi-percentage-point range on the {\tt swe\_bench}
family, while under~(b) the same benchmarks still shift by several
points. Two observations follow. First, even the conservative
Major-only filter moves leaderboard numbers enough to matter for
ranking decisions, which directly motivates our recommendation to
publish audit-aware scores alongside raw pass rates. Second, the
gap between~(a) and~(b) shows that Minor findings are not
cosmetic: the share of measured performance attributable to
ambiguous or partially-broken tasks is a non-trivial fraction of
the headline score on every benchmark in this slice.

\subsection{Trajectory audit independence from agent outcome}
\label{app:outcome_matrix}

\begin{figure}[t]
  \centering
  \includegraphics[width=\linewidth]{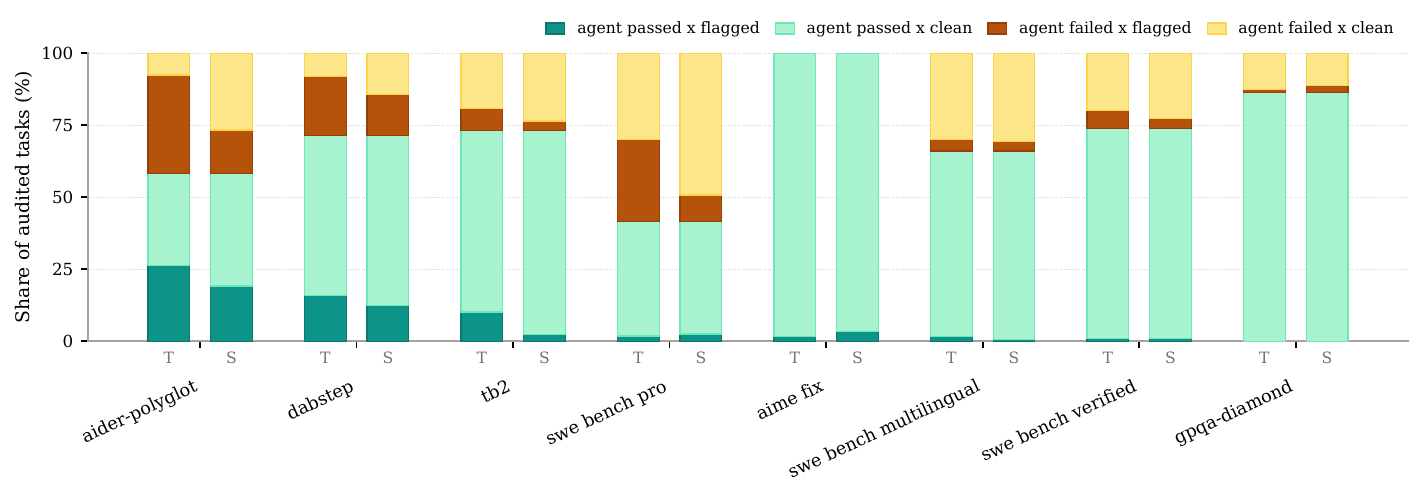}
  \caption{Audit flag $\times$ agent outcome at max severity on
  the paired-mode benchmarks. For each benchmark we show two
  100\,\%-normalized stacked bars -- trajectory (T) and static
  (S) -- segmented into the four cells of the audit-flag $\times$
  agent-outcome matrix at the rubric's max severity:
  \emph{agent passed $\times$ flagged}, \emph{agent passed $\times$
  clean}, \emph{agent failed $\times$ flagged}, and
  \emph{agent failed $\times$ clean}. If trajectory flags merely
  tracked agent failure, the dark (flagged) slices on the T bar
  would concentrate in the agent-failed half; in practice the T
  bar carries a comparable or larger \emph{agent passed $\times$
  flagged} slice than the S bar on most paired benchmarks.}
  \label{fig:outcome_matrix}
\end{figure}

A natural objection to the leaderboard lifts in
\S\ref{sec:ablations} is that the trajectory auditor flags
more (and more severely) than the static auditor only because it
sees which tasks the agent failed -- in which case its findings
would be a re-skin of the score it is supposed to audit. We test
this directly by conditioning on agent outcome.
Figure~\ref{fig:outcome_matrix} reports, on the paired-mode
benchmarks, the
share of audited tasks in each cell of the audit-flag $\times$
agent-outcome matrix at the rubric's max severity, separately for
the trajectory (T) and static (S) audits. The bias-only hypothesis
predicts that trajectory-flagged tasks collapse into the
\emph{agent-failed} half whenever trajectory's overall flag count
exceeds static's. The pattern deviates from that prediction on
most paired benchmarks: trajectory's \emph{agent-passed $\times$
flagged} slice is non-trivial and at times exceeds static's,
indicating that trajectory's extra max-severity findings are not
concentrated in agent-failed tasks alone.

\subsection{Leaderboard shifts}
\label{app:leaderboards:shifts}

\begin{figure}[t]
  \centering
  \includegraphics[width=0.95\linewidth]{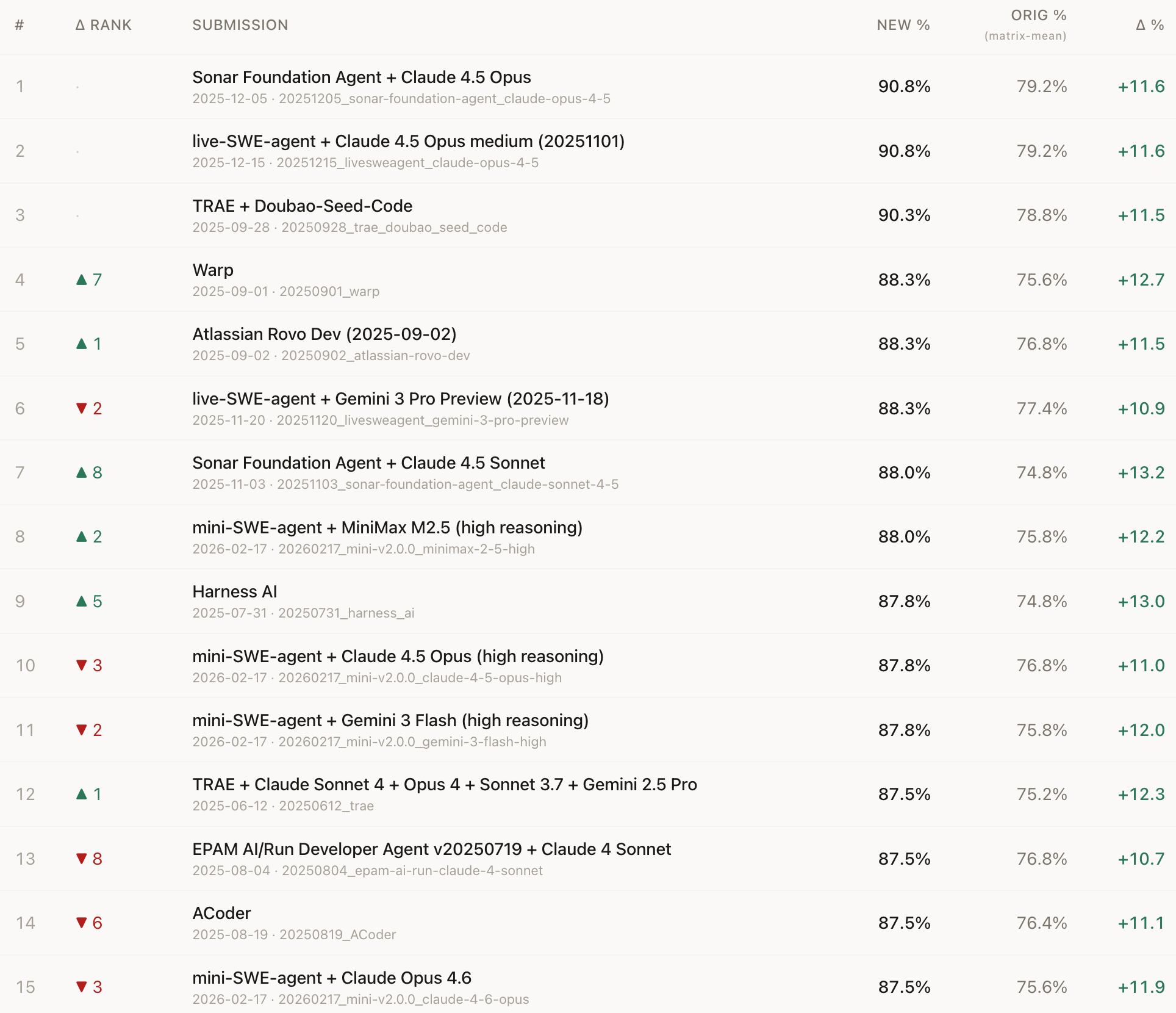}
  \caption{Top-15 of {\tt swe\_bench\_verified} after removing all
  Major and Minor trajectory-flagged tasks. \textsc{orig~\%} is the
  public matrix-mean score; \textsc{new~\%} is recomputed on the
  audited task set; $\Delta$~\% and $\Delta$~Rank report the change
  relative to the public ordering. Even with the broader exclusion
  rule the top three slots are stable, but several middle-of-pack
  systems swing by up to eight places (e.g.\ Sonar Foundation Agent
  + Claude 4.5 Sonnet rises eight, EPAM AI/Run Developer Agent
  drops eight), with the largest score lifts exceeding +13 points.}
  \label{fig:leaderboard:swebv:majmin}
\end{figure}

\begin{figure}[t]
  \centering
  \includegraphics[width=0.95\linewidth]{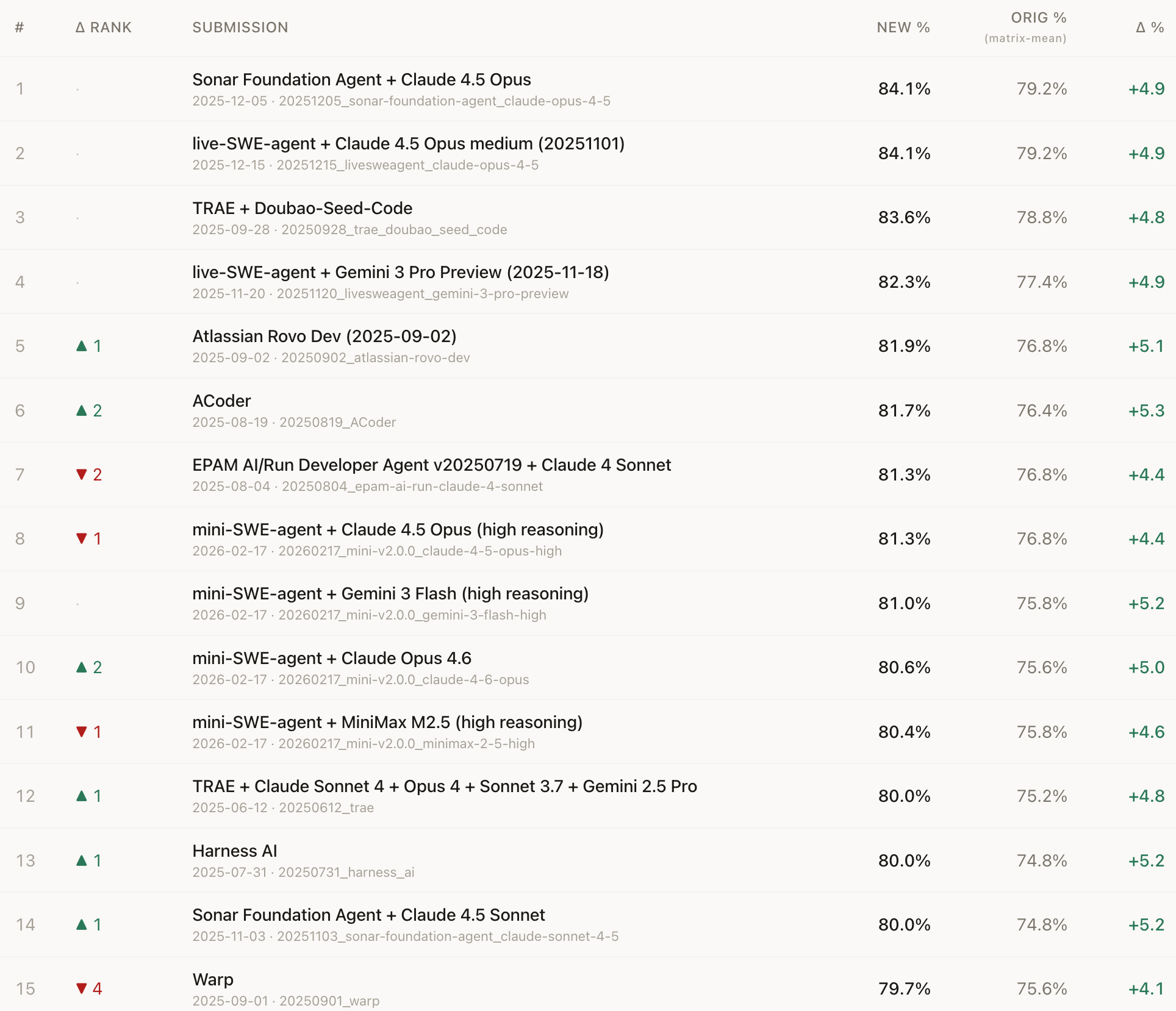}
  \caption{Top-15 of {\tt swe\_bench\_verified} after removing only
  Major trajectory-flagged tasks (the conservative filter). The
  rank reshuffle is smaller than under the Major+Minor filter --
  most movements are within $\pm 2$ -- but the average per-model
  lift remains around five points, and the head of the leaderboard
  still re-orders around the 5th--14th slots.}
  \label{fig:leaderboard:swebv:maj}
\end{figure}

\begin{figure}[t]
  \centering
  \includegraphics[width=0.95\linewidth]{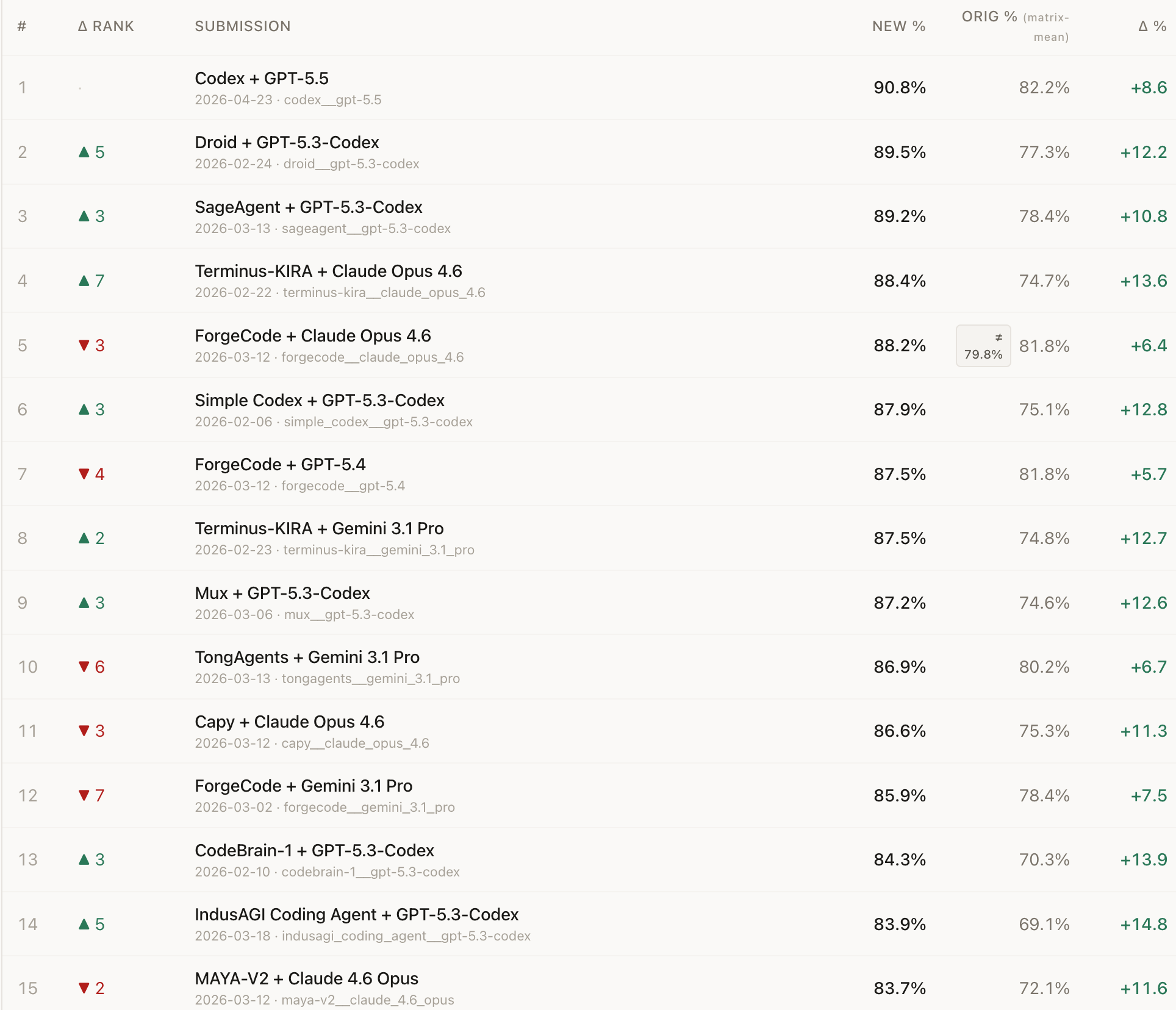}
  \caption{Top-15 of Terminal-Bench~2 after removing all Major and
  Minor trajectory-flagged tasks. Terminal-Bench~2 reshuffles more
  aggressively than {\tt swe\_bench\_verified}: a non-trivial share
  of the public task set is flagged, lifting per-model scores by
  several percentage points and moving multiple submissions by
  three or more rank positions.}
  \label{fig:leaderboard:tb2:majmin}
\end{figure}

\begin{figure}[t]
  \centering
  \includegraphics[width=0.95\linewidth]{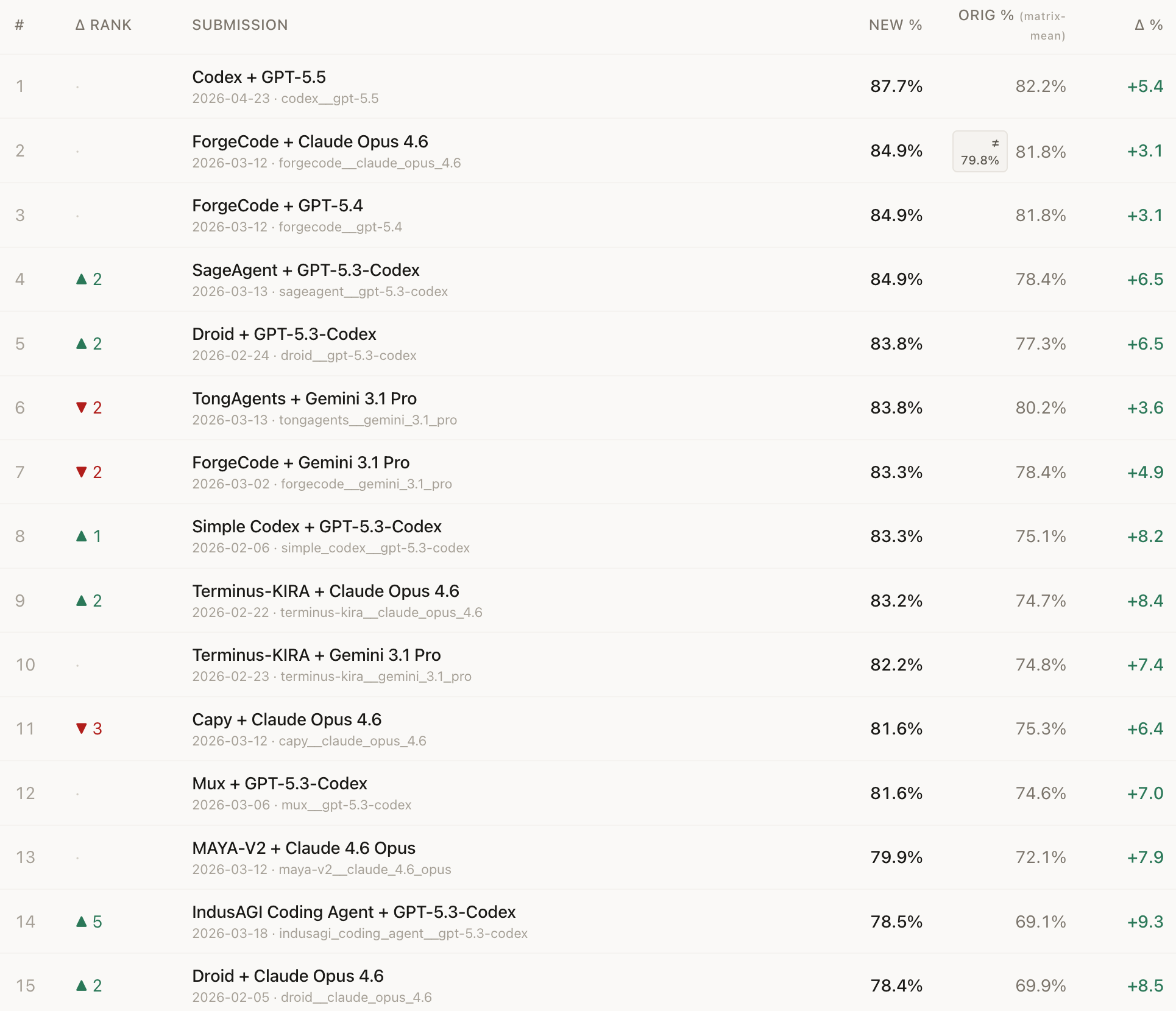}
  \caption{Top-15 of Terminal-Bench~2 after removing only Major
  trajectory-flagged tasks. Even under this conservative filter,
  Terminal-Bench~2 sees more rank movement than {\tt swe\_bench\_verified}
  under the same rule, indicating that a meaningful fraction of the
  public score is attributable to tasks the trajectory audit flags
  at the Major bar alone.}
  \label{fig:leaderboard:tb2:maj}
\end{figure}

The mean per-model deltas in
\S\ref{app:leaderboards:meandelta} aggregate across submissions, but
individual models can move in different directions and by different
amounts when flagged tasks are removed -- so the headline lift does
not, by itself, indicate how much published rankings reshuffle. To
make the shifts concrete, we recompute the public leaderboards on
two representative benchmarks, swe\_bench\_verified and
Terminal-Bench~2, after excluding the audit-flagged tasks. All
findings are taken from the trajectory audit, and all submissions
shown are unchanged from their public entries; only the underlying
task set differs. For each benchmark we present two snapshots: a
strict variant that drops every Major and Minor flagged
task, and a conservative variant that drops only Major flagged
tasks. Each snapshot reports the original public score
(\textsc{orig~\%}, computed as the matrix-mean over the unrevised
task set), the recomputed score on the audited task set
(\textsc{new~\%}), the per-model score change ($\Delta$~\%), and
the rank change relative to the public ordering ($\Delta$~Rank).
Figures~\ref{fig:leaderboard:swebv:majmin}--\ref{fig:leaderboard:tb2:maj}
show the top-15 entries on each combination.

\clearpage

\section{Detailed audit case studies}
\label{app:case_studies}

This appendix expands four findings from the audit into the same
\textsc{Prompt} / \textsc{Evaluation} / \textsc{Finding} layout
used in Figure~\ref{fig:hook}, with additional context: the
rubric category, severity, benchmark domain, and the human
reviewer's verdict at the consolidation step
(\S\ref{sec:audit:validation:manual}). The four cases span all
three rubric categories, three benchmarks, and the three
reviewer outcomes (\emph{Aligned}, \emph{Partial},
\emph{Disagree}); finding text is excerpted verbatim from the
per-task audit JSON files released with the artifact.

\newcommand{\auditchip}[2]{%
  \tcbox[on line, nobeforeafter, colback=#1!10, colframe=#1!75!black,
    boxrule=0.4pt, arc=2pt, left=2.5pt, right=2.5pt, top=0pt, bottom=0pt,
    boxsep=0.5pt, fontupper=\scriptsize\sffamily]{#2}\hspace{1.5pt}%
}
\newcommand{\auditverdict}[1]{%
  \tcbox[on line, nobeforeafter, colback=black!5, colframe=black!55,
    boxrule=0.4pt, arc=2pt, left=2.5pt, right=2.5pt, top=0pt, bottom=0pt,
    boxsep=0.5pt, fontupper=\scriptsize\sffamily]{Human review:~\textbf{#1}}%
}
\newcommand{\auditsection}[3]{%
  \begin{tcolorbox}[
    enhanced, width=\linewidth,
    colback=#1!50, opacityback=0.16,
    frame hidden, boxrule=0pt, arc=1.5pt,
    boxsep=0pt, left=4pt, right=4pt, top=2pt, bottom=2pt,
    fontupper=\footnotesize,
  ]
    \textcolor{#1!70!black}{\textbf{#2}}~#3
  \end{tcolorbox}%
}

\subsection{Instruction, Major: \texttt{install-windows-3.11}
(Terminal-Bench~2)}
\label{app:case:install_windows}

\begin{tcolorbox}[
  enhanced, breakable, width=\linewidth,
  colback=white, colframe=rubricA!70!black,
  colbacktitle=rubricA!80!black, coltitle=white,
  boxrule=0.5pt, arc=2pt,
  left=4pt, right=4pt, top=3pt, bottom=3pt,
  toptitle=2pt, bottomtitle=2pt,
  fonttitle=\bfseries\small,
  title={\textsc{Instruction}\hfill
    \normalfont\sffamily\footnotesize Terminal-Bench~2~/
    \texttt{install-windows-3.11}},
]
\auditchip{rubricA}{Category: Instruction}%
\auditchip{rubricA}{Severity: Major}%
\auditchip{rubricA}{Domain: Coding}%
\auditverdict{Aligned}

\medskip
\auditsection{rubricA}{Prompt.}{%
``Configure QEMU to accept keyboard input programmatically~\dots\
Set up QEMU with appropriate interfaces to allow external
keyboard control.'' (\texttt{instruction.md:10--12}). The prompt
specifies neither the monitor protocol (HMP vs.\ QMP), the
transport (TCP, telnet, stdio, UNIX socket), nor any path.
}\vspace{-7pt}

\auditsection{rubricA}{Evaluation.}{%
\texttt{tests/test\_outputs.py} hardcodes
\texttt{monitor\_socket = "/tmp/qemu-monitor.sock"} and runs
\texttt{socat - UNIX-CONNECT:\{monitor\_socket\}} while piping
HMP commands such as \texttt{sendkey f1}. Only a UNIX-socket HMP
monitor at exactly this path passes
\texttt{test\_windows\_keys\_with\_visual\_feedback}.
}\vspace{-7pt}

\auditsection{rubricA}{Finding.}{%
The visual-feedback keyboard test requires a QEMU HMP monitor
exposed as a UNIX socket at \texttt{/tmp/qemu-monitor.sock}, but
the prompt never discloses the path, transport, or protocol.
QEMU satisfies the stated requirement with many valid
configurations -- HMP over TCP, telnet, stdio, a UNIX socket at
a different path, or QMP over UNIX/TCP. The reference
solution uses
\texttt{-monitor unix:/tmp/qemu-monitor.sock,server,nowait}
(\texttt{solution/solve.sh:118}), the exact path/protocol
required by the test, but nothing in the prompt points an agent
to this specific choice.
}\vspace{-7pt}

\auditsection{rubricA}{Why it matters.}{%
A competent sysadmin reading only the prompt has no basis to
converge on this exact filename. The fix is either to disclose
the contract (``the VM's HMP monitor must be exposed at
\texttt{/tmp/qemu-monitor.sock} as a UNIX socket'') or to relax
the test to discover any QEMU monitor interface from the
command line.
}
\end{tcolorbox}

\subsection{Evaluation, Major: \texttt{validation\_Design\_15}
(MMMU-Pro)}
\label{app:case:design_15}

\begin{tcolorbox}[
  enhanced, breakable, width=\linewidth,
  colback=white, colframe=rubricT!70!black,
  colbacktitle=rubricT!80!black, coltitle=white,
  boxrule=0.5pt, arc=2pt,
  left=4pt, right=4pt, top=3pt, bottom=3pt,
  toptitle=2pt, bottomtitle=2pt,
  fonttitle=\bfseries\small,
  title={\textsc{Evaluation}\hfill
    \normalfont\sffamily\footnotesize MMMU-Pro~/
    \texttt{validation\_Design\_15}},
]
\auditchip{rubricT}{Category: Evaluation}%
\auditchip{rubricT}{Severity: Major}%
\auditchip{rubricT}{Domain: Multimodal}%
\auditverdict{Aligned}

\medskip
\begin{center}
  \includegraphics[width=0.62\linewidth]{figures/audit_example_design15.png}\\[2pt]
  {\scriptsize\itshape Reference image associated with the task
  (\texttt{std\_image\_1.png}).}
\end{center}

\auditsection{rubricT}{Prompt.}{%
``The work (\textlangle image~1\textrangle) on the left was
made for a~\dots'' Standard~(4 options): A.~royal palace,
B.~religious institution, C.~private home, D.~civic building.
The 10-option and vision configs ask the same question with
the answer text \emph{royal palace} preserved across configs.
}\vspace{-7pt}

\auditsection{rubricT}{Evaluation.}{%
\texttt{evaluate.py:14--31} (\texttt{mmmu\_process\_results})
parses the response with
\texttt{parse\_multi\_choice\_response()} and compares the
parsed letter against the gold \texttt{answer} field via exact
string equality. The gold answer is~``A''~(4-option) /~``C''
~(10-option, vision), all mapping to ``royal palace''. There is
no allowance for alternative correct answers.
}\vspace{-7pt}

\auditsection{rubricT}{Finding.}{%
The gold answer ``royal palace'' is factually wrong for this
image. The work depicted is Fra Angelico's \emph{Annunciation}
(San~Marco corridor fresco, c.~1440--1445), commissioned for
the Dominican Convent of San~Marco in Florence -- a religious
institution. None of Fra Angelico's major
\emph{Annunciations} (San~Marco, Cortona, Prado/Fiesole) were
made for a royal palace; all were made for Dominican religious
institutions. The same wrong key propagates across the
4-option, 10-option, and vision configs because they share the
underlying answer key.
}\vspace{-7pt}

\auditsection{rubricT}{Why it matters.}{%
A test-taker who correctly identifies this canonical work would
answer ``religious institution'' and be marked incorrect. The
defect is not a discoverable gap; no standard tool or reasoning
path bridges to ``royal palace'' for a Fra Angelico
\emph{Annunciation}. The test penalizes accurate
art-historical knowledge -- the inverse of its stated purpose.
}
\end{tcolorbox}

\subsection{Evaluation, Major: \texttt{cvpr-research}
(Toolathlon)}
\label{app:case:cvpr_research}

\begin{tcolorbox}[
  enhanced, breakable, width=\linewidth,
  colback=white, colframe=rubricT!70!black,
  colbacktitle=rubricT!80!black, coltitle=white,
  boxrule=0.5pt, arc=2pt,
  left=4pt, right=4pt, top=3pt, bottom=3pt,
  toptitle=2pt, bottomtitle=2pt,
  fonttitle=\bfseries\small,
  title={\textsc{Evaluation}\hfill
    \normalfont\sffamily\footnotesize Toolathlon~/
    \texttt{cvpr-research}},
]
\auditchip{rubricT}{Category: Evaluation}%
\auditchip{rubricT}{Severity: Major}%
\auditchip{rubricT}{Domain: Interactive / Agent / Tool Use}%
\auditverdict{Partial}

\medskip
\auditsection{rubricT}{Prompt.}{%
List the three CVPR~2025 researchers who simultaneously
(a)~publish the most at the venue and (b)~match the user's
profile from \texttt{personal\_info.md}: a postdoc seeker in
Hong~Kong, working on visual generative models
(diffusion / flow matching). The output must be three lines, one
researcher per line.
}\vspace{-7pt}

\auditsection{rubricT}{Evaluation.}{%
\texttt{evaluation/main.py:29--42} loops over the three lines
and checks whether \emph{any} line contains the normalized
substring \texttt{leizhang}/\texttt{zhanglei} and whether
\emph{any} line contains
\texttt{hongshengli}/\texttt{lihongsheng}; the third slot is
never checked. Normalization
(\texttt{utils/general/helper.py:547--549}) strips all non-word
characters and lowercases.
}\vspace{-7pt}

\auditsection{rubricT}{Finding.}{%
The evaluation only verifies two of the three required names,
and even those two checks use coarse substring matching on
extremely common Chinese names. ``Lei~Zhang'' at any institution
matches; the prompt's affiliation (Hong~Kong) and topic
(visual generative models) constraints are not checked. The
authors themselves acknowledge the third entry is unverifiable
in a comment in \texttt{main.py:29--31}: ``haochen is not
detected, as the statistics from paper copilot may be
inaccurate.''
}\vspace{-7pt}

\auditsection{rubricT}{Why it matters.}{%
A solution that names two unrelated Lei~Zhang / Hongsheng~Li
researchers plus a random third string passes, while a
correct, semantically-aligned answer that uses different (but
equally valid) surname spellings might not. The test does not
evaluate what the prompt asks; tightening the eval to use
affiliation-aware lookups and a frozen CVPR~2025 author table
would resolve both halves of the defect.
}
\end{tcolorbox}

\subsection{Evaluation, Major: \texttt{vlm-history-completer}
(Toolathlon)}
\label{app:case:vlm_history}

\begin{tcolorbox}[
  enhanced, breakable, width=\linewidth,
  colback=white, colframe=rubricT!70!black,
  colbacktitle=rubricT!80!black, coltitle=white,
  boxrule=0.5pt, arc=2pt,
  left=4pt, right=4pt, top=3pt, bottom=3pt,
  toptitle=2pt, bottomtitle=2pt,
  fonttitle=\bfseries\small,
  title={\textsc{Evaluation}\hfill
    \normalfont\sffamily\footnotesize Toolathlon~/
    \texttt{vlm-history-completer}},
]
\auditchip{rubricT}{Category: Evaluation}%
\auditchip{rubricT}{Severity: Major}%
\auditchip{rubricT}{Domain: Interactive / Agent / Tool Use}%
\auditverdict{Disagree}

\medskip
\auditsection{rubricT}{Prompt.}{%
Populate the Architecture and Sources columns of a Google~Sheet
for 20 vision--language models. \texttt{Order.txt} states an
explicit priority rule for source URLs: \emph{official websites
or releases or news or blogs} $>$ \emph{official GitHub
repository} $>$ \emph{ArXiv paper}, and ``if higher priority
sources are available, filling in the lower priority sources
will not be accepted.''
}\vspace{-7pt}

\auditsection{rubricT}{Evaluation.}{%
\texttt{evaluation/main.py:172--190} scores Sources by checking
whether any groundtruth URL (after stripping non-word
characters) appears as a substring of the submitted cell.
Lines~274--278 require an overall score $\geq 1.0$ -- a 100\%
pass threshold over a hand-picked URL whitelist, with no
priority logic.
}\vspace{-7pt}

\auditsection{rubricT}{Finding.}{%
The Stable Diffusion~1.5 reference source in
\texttt{groundtruth.json:47--51} is
\texttt{https://github.com/Kameronski/stable-diffusion-1.5}, a
personal user fork. This is not an ``official GitHub
repository'' as \texttt{Order.txt} requires. A compliant agent
will instead find Stability~AI's news/blog (highest priority)
or an official repo (e.g., \texttt{CompVis/stable-diffusion},
\texttt{runwayml/stable-diffusion-v1-5}); none of these are
accepted by the substring-whitelist eval.
}\vspace{-7pt}

\auditsection{rubricT}{Why it matters.}{%
An agent that correctly follows the stated priority rule cannot
pass this entry, because the only accepted URL is a low-priority
personal fork that violates that rule. The only way to pass is
to know the specific arbitrary URL the benchmark picked --
which is not discoverable from the prompt or standard domain
knowledge. The human reviewer disagreed with this finding;
we keep the case here as an example of a flagged item that did
not survive consolidation.
}
\end{tcolorbox}

\clearpage
\section{Limitations}
\label{app:limitations}

ABA is a useful tool, not an oracle. The auditor is itself an LLM
(Claude Opus~4.7) and inherits its failure modes, our validation
covers only a sampled fraction of the 34{,}285 audited tasks
(\S\ref{sec:audit:validation}), and the severity rubric
(\S\ref{app:rubrics:severity}) is interpretive rather than
mechanical, so reasonable auditors will disagree on borderline
cases. Findings also reflect a snapshot at fixed commits and
exclude benchmark types outside our portfolio (e.g., audio,
embodied 3D, subjective-evaluation tasks), so domain-level rates
and any downstream ranking shifts should be read as ABA's
estimates relative to the versions audited rather than as ground
truth.

\section{Ethics and data considerations}
\label{app:ethics}

\paragraph{Code of ethics.} This work conforms to the Code
of Ethics. The study audits the design of public benchmark
artifacts (task prompts, environments, evaluation code, and
metadata); it involves no human subjects, no crowdsourced
labelling, and no collection of personal data. The auditor itself
is an off-the-shelf LLM coding agent
(\S\ref{app:reproducibility}), and all model invocations stay
within the API provider's published usage policy. Audit findings
are framed constructively: they identify task-level defects
(specification gaps, environment fragility, brittle graders) and
are intended to help benchmark authors fix issues, not to
penalise individual benchmarks or the models evaluated on them.
Where our findings reproduce small excerpts of benchmark prompts
or grader code (e.g., the case studies in
\S\ref{app:case_studies}), we limit excerpts to what is needed to
make the finding interpretable and cite the upstream source. The
released artifacts include no pre-trained model, generative
system, or scraped dataset, so they do not pose the misuse risks
that typically motivate gated release; the residual risk specific
to this work is that flagged tasks could be selectively excluded
to inflate reported scores. We mitigate this by releasing every
finding with its evidence pointers (task identifier, file/line
spans, and rubric calibration) so any third party can
independently verify or contest a finding, and by reporting
leaderboard sensitivity under explicit exclusion rules
(\S\ref{app:leaderboards:meandelta}) rather than publishing a
curated ``cleaned'' leaderboard.

\paragraph{Data provenance.} The audited portfolio is assembled
from two public sources documented in
\S\ref{app:portfolio}: (i)~benchmarks named in the headline
capability tables of five public frontier model release reports,
and (ii)~accepted papers from the NeurIPS~2025 Datasets~\&~Benchmarks
Track. Each audited benchmark is cited to its upstream release
and used under that release's published license and terms of use;
we do not modify or redistribute the benchmark tasks themselves.
Per-benchmark provenance -- the canonical name, upstream
citation, version or commit pin, and audited subset -- is recorded
in the per-benchmark inventory
(\S\ref{app:portfolio:inventory}). No benchmark in the portfolio
is private, gated, or sourced from human subjects collected by us.

\paragraph{Release plan.} The artifacts released with this paper
cover the auditor and its outputs, not the upstream benchmark
data. Concretely, we release: the evidence-collector manifests
and run scripts, the auditor prompts and severity rubric
(\S\ref{app:prompts}), the Claude Code CLI configuration
(\S\ref{app:reproducibility}), the per-task audit findings with
their evidence pointers, and the scoring and aggregation scripts
used to produce every table and figure in the paper. Because the
audited benchmarks span many different upstream licenses, we do
not redistribute their task content under a single license;
instead, the released metadata points back to each upstream
release so downstream users can obtain the underlying tasks under
the upstream terms.

\clearpage

\end{document}